\documentclass[runningheads]{llncs}

\usepackage{eccv}

\usepackage{eccvabbrv}
\definecolor{cvprblue}{rgb}{0.21,0.49,0.74}
\usepackage[pagebackref,breaklinks]{hyperref}
\usepackage{colortbl}
\usepackage[table]{xcolor}

\usepackage{graphicx}
\usepackage{booktabs}
\usepackage{verbatim}
\usepackage{listings}
\usepackage{multirow}

\usepackage[accsupp]{axessibility}  

\usepackage{orcidlink}

\usepackage{graphicx}
\usepackage{tabularx}
\usepackage{booktabs}
\usepackage{multirow}

\usepackage[accsupp]{axessibility}  

\usepackage{hyperref}
\usepackage{orcidlink}

\begin{document}

\title{From Unlearning to UNBRANDING: A Benchmark for Trademark-Safe Text-to-Image Generation} 

\titlerunning{UnBranding}


\author{
Dawid Malarz\textsuperscript{*1,3}\quad
Filip Manjak\textsuperscript{*1}\quad
Maciej Zieba\textsuperscript{2}\\[5pt]
Przemys\l{}aw Spurek\textsuperscript{1,3}
Artur Kasymov\textsuperscript{1}\quad
}

\authorrunning{Malarz. et al.}


\institute{Jagiellonian University \and
Wroc\l{}aw University of Science and Technology \and
IDEAS Research Institute\\
\email{przemyslaw.spurek@uj.edu.pl}\\
\textsuperscript{*} Equal contribution}

\maketitle

\vspace{-0.25cm}
\begin{figure}
    \begin{center}
    \vspace{-0.4cm}
    \includegraphics[width=0.99\textwidth]{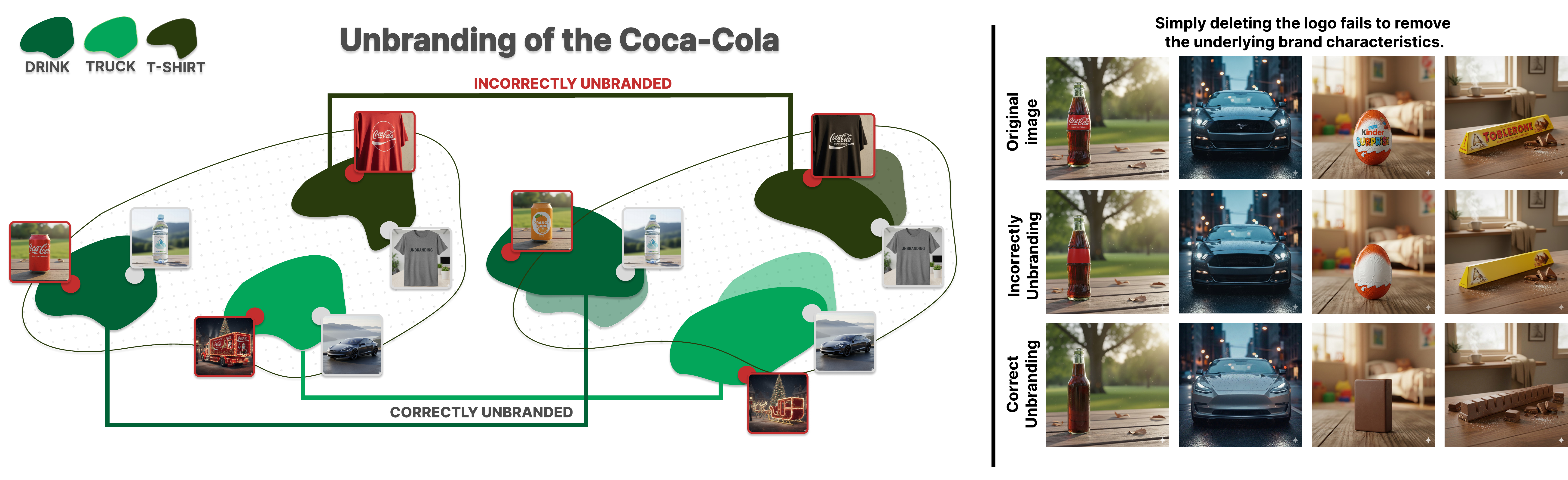}
    \vspace{-0.4cm}
    \end{center}
    \caption{ 
    Illustration of the {\bf unbranding} task using the Coca-Cola brand. Coca-Cola identifiers appear across diverse contexts, including clothing, beverages, and vehicles, requiring brand removal to operate across multiple object types rather than on a single concept as in classical unlearning. Effective unbranding must modify all brand-bearing regions and address more than logo removal, since brand identity can also arise from distinctive trade dress features such as the iconic Coca-Cola bottle shape.
    }
    \label{fig:teser}
    \vspace{-1.4cm}
\end{figure}

\begin{abstract}
The rapid progress of text-to-image diffusion models raises significant concerns regarding the unauthorized reproduction of trademarked content. While prior work targets general concepts (e.g., styles, celebrities), it fails to address specific brand identifiers. Brand recognition is multi-dimensional, extending beyond explicit logos to encompass distinctive structural features (e.g., a car's front grille). To tackle this, we introduce {\bf unbranding}, a novel task for the fine-grained removal of both trademarks and subtle structural brand features, while preserving semantic coherence. We construct a benchmark dataset and introduce a novel evaluation framework combining Vision Language Models (VLMs) with segmentation-based classifiers trained on human annotations of logos and trade dress features, addressing the limitations of existing brand detectors that fail to capture abstract trade dress. Furthermore, we observe that newer, higher-fidelity systems (SDXL, FLUX) synthesize brand identifiers more readily than older models, highlighting the urgency of this challenge. Our results confirm that unbranding is a distinct problem requiring specialized techniques.

\end{abstract}

\section{Introduction}

Recent advances in text-to-image diffusion models \cite{ho2020denoisingdiffusionprobabilisticmodels, ramesh2022hierarchicaltextconditionalimagegeneration, saharia2022photorealistictexttoimagediffusionmodels} have unlocked the ability to synthesize images with remarkable fidelity, capturing fine textures, realistic objects, and complex compositions. These capabilities are driving an explosion of creative applications, but they also surface new risks. In particular, generative models often reproduce trademarked logos, branded objects, and distinctive design elements without authorization. Such outputs are not only problematic from an intellectual property perspective (e.g., trademark dilution, consumer confusion) but also represent a barrier to safe deployment of generative models in real-world, commercial, and high-stakes domains.

While prior work has analyzed brand signals in machine learning \cite{qraitem2024slantspuriouslogoanalysis,kiapour2018brandlogovisual} or explored coarse-grained concept erasure (unlearning) \cite{gandikota2023erasing, kumari2023ablating}, these approaches are ill-suited for the generative removal of brand identifiers. Existing methods either focus on detection, not removal, or are too coarse, failing to preserve the underlying object's integrity when an entire concept is erased. The increasing fidelity of modern generative models magnifies this challenge. We observe that newer systems (e.g., SDXL~\cite{podell2023sdxl}, FLUX~\cite{batifol2025flux}) reproduce trademarks with far greater accuracy than previous systems \cite{ho2020denoisingdiffusionprobabilisticmodels}, thereby increasing the risk of unauthorized reproduction. Also, LMMs (Large Multimodal Models), such as Gemini or ChatGPT, produce high-quality brand names that can be readily used in commercial applications.  Thus, across both the unlearning and logo-analysis literatures, a striking gap remains: no existing method enables a generative model to retain an object while reliably erasing its branding.

\begin{figure}[]
  \centering
  \includegraphics[width=0.24\linewidth]{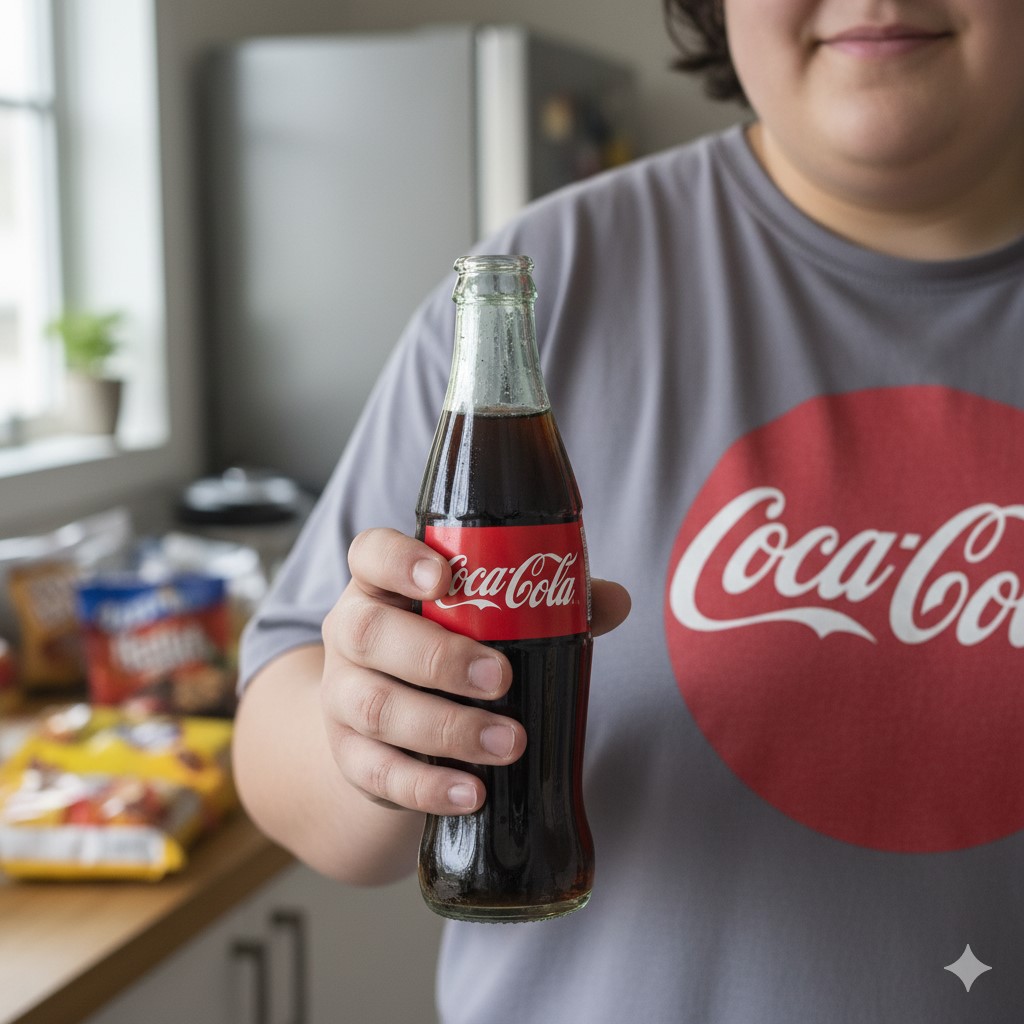}
  \includegraphics[width=0.24\linewidth]{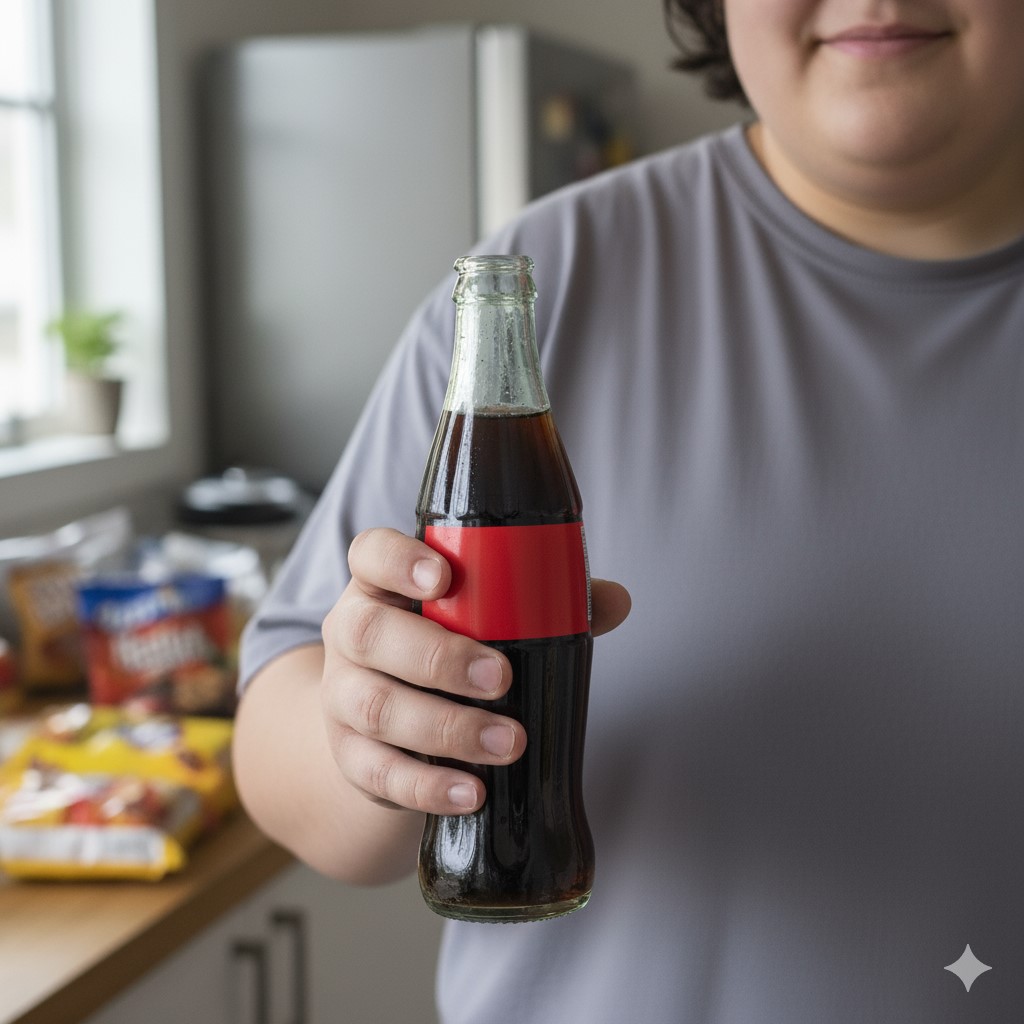}
 \includegraphics[width=0.24\linewidth]{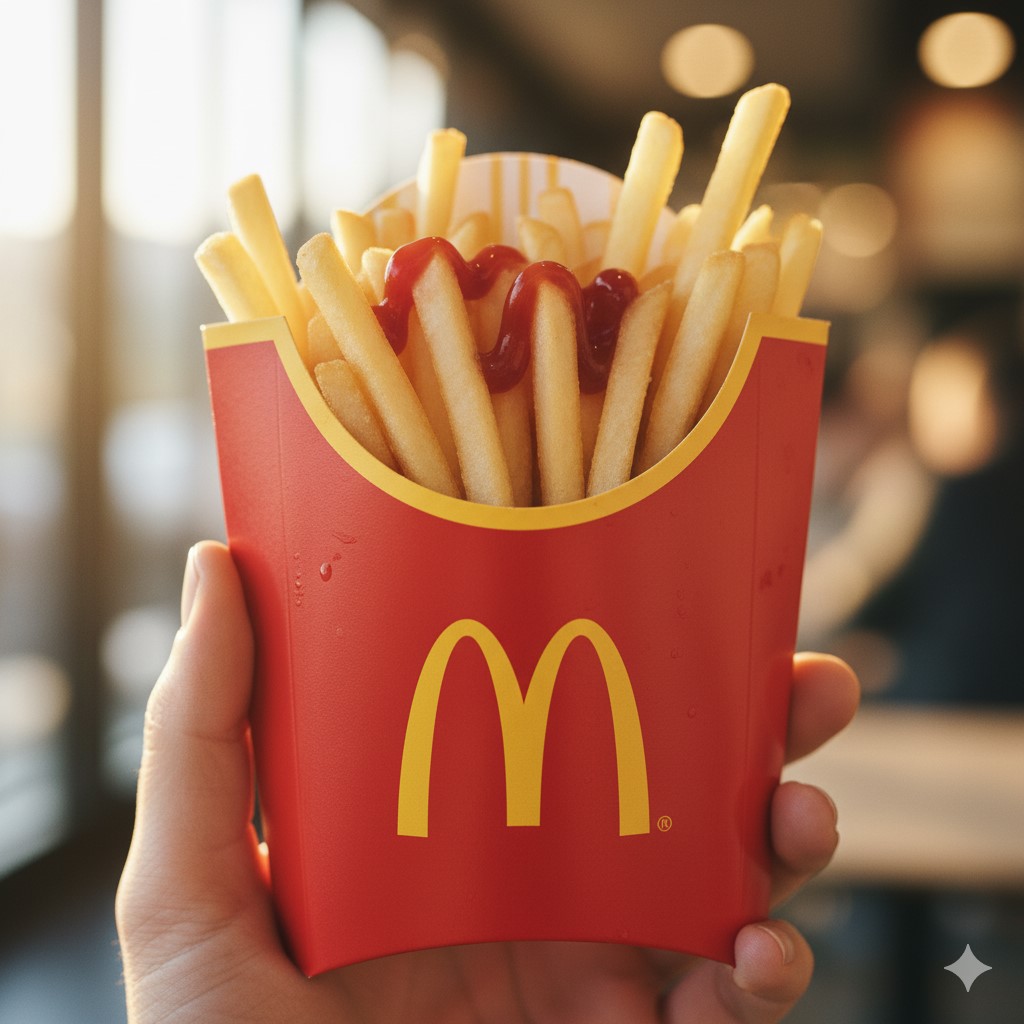}
  \includegraphics[width=0.24\linewidth]{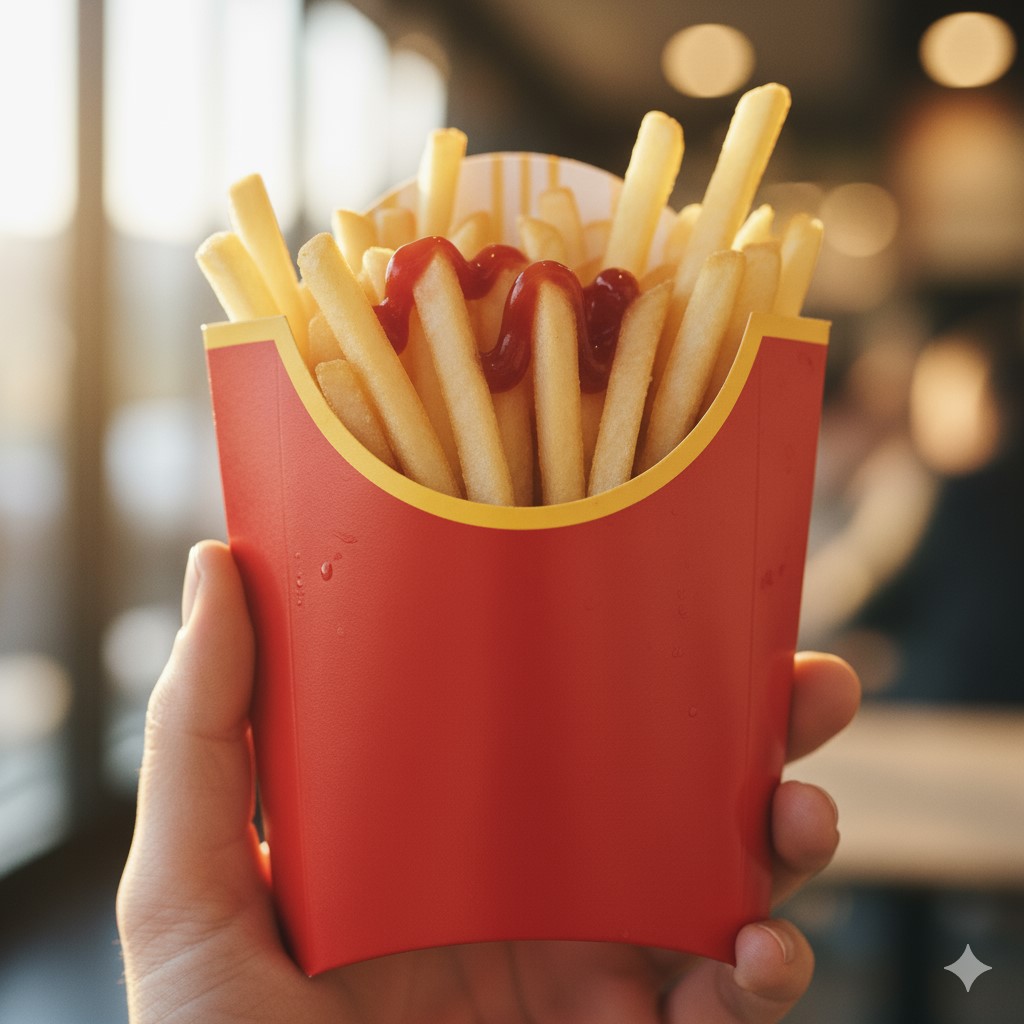}
\\
\includegraphics[width=0.24\linewidth]{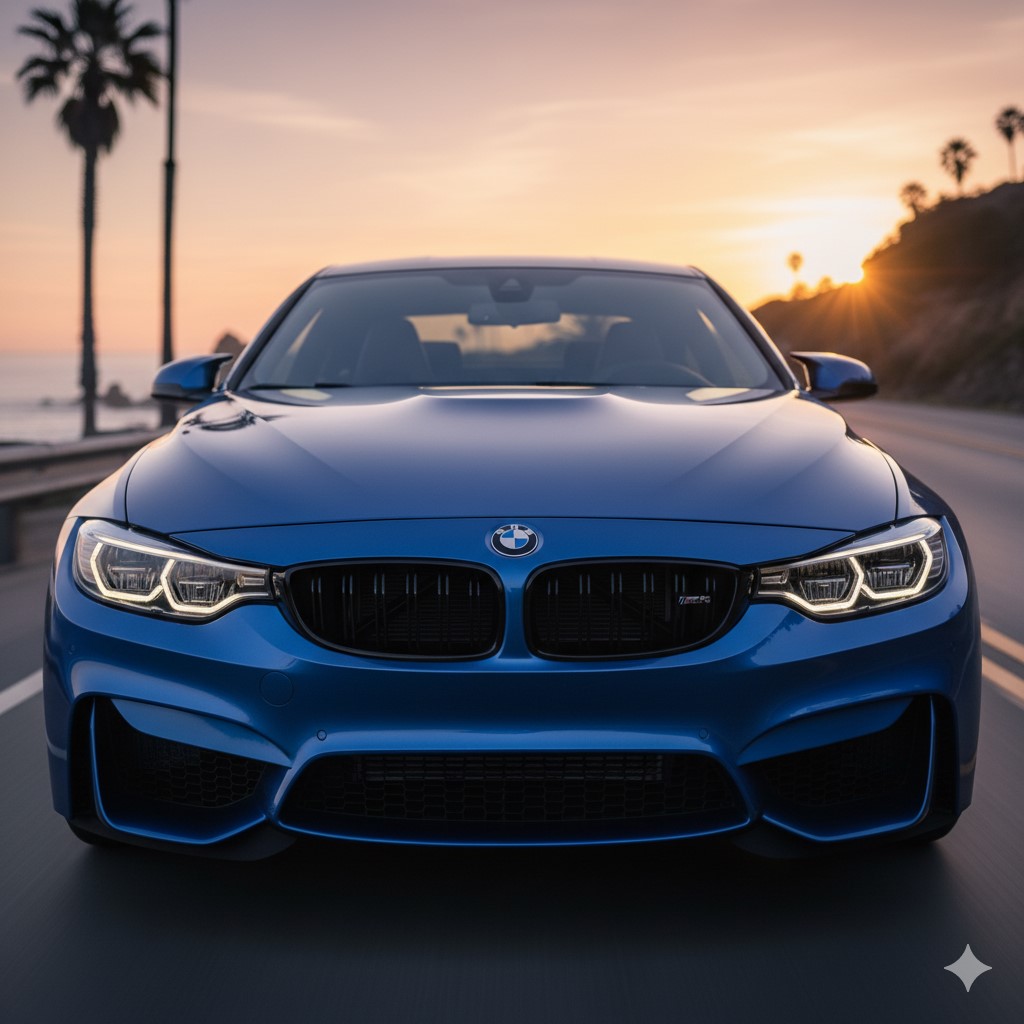}
  \includegraphics[width=0.24\linewidth]{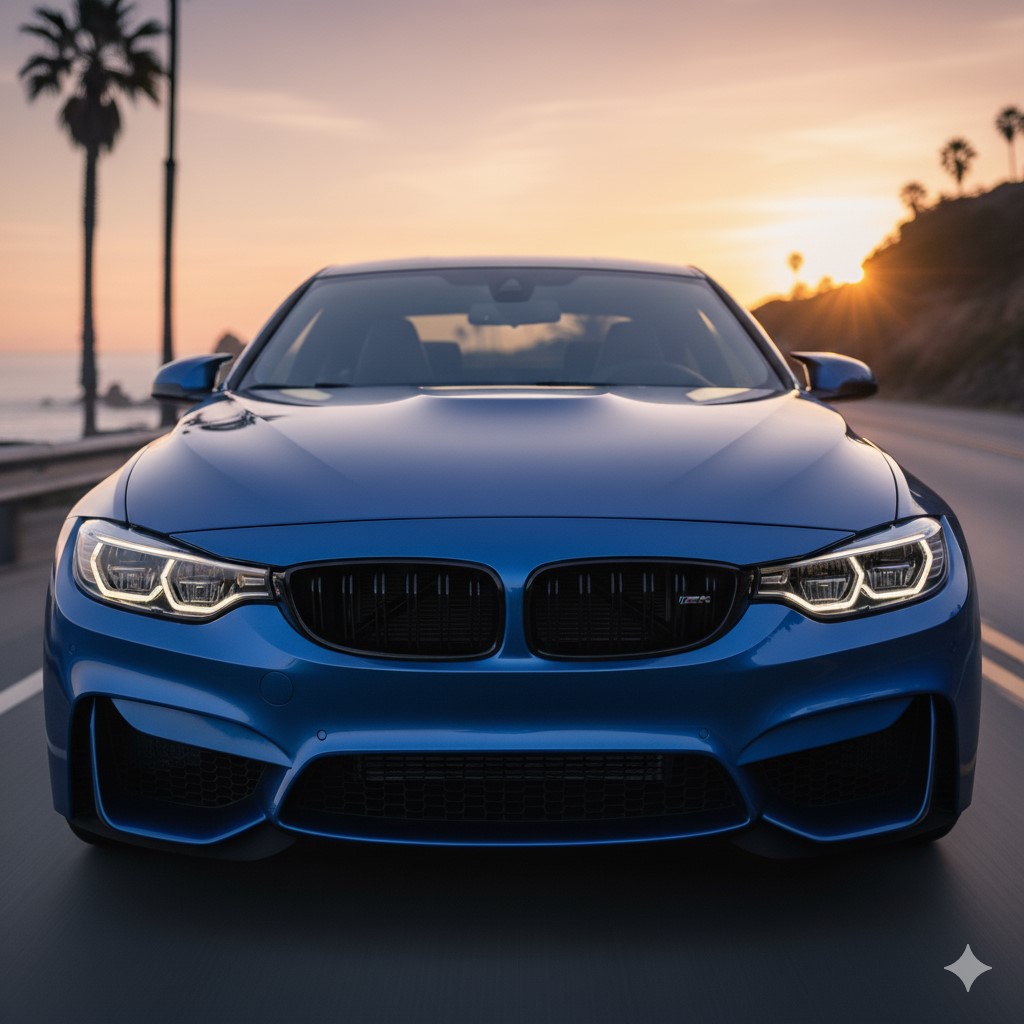}
\includegraphics[width=0.24\linewidth]{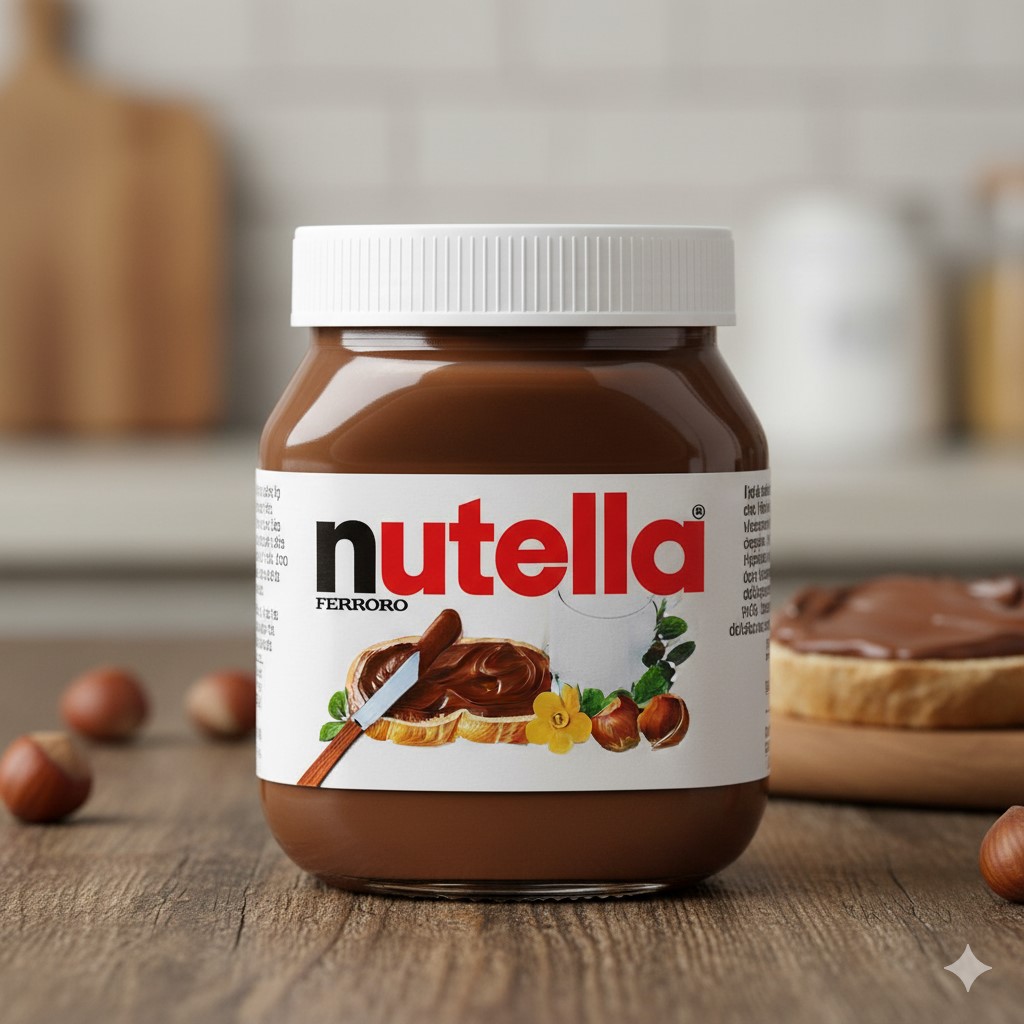}
  \includegraphics[width=0.24\linewidth]{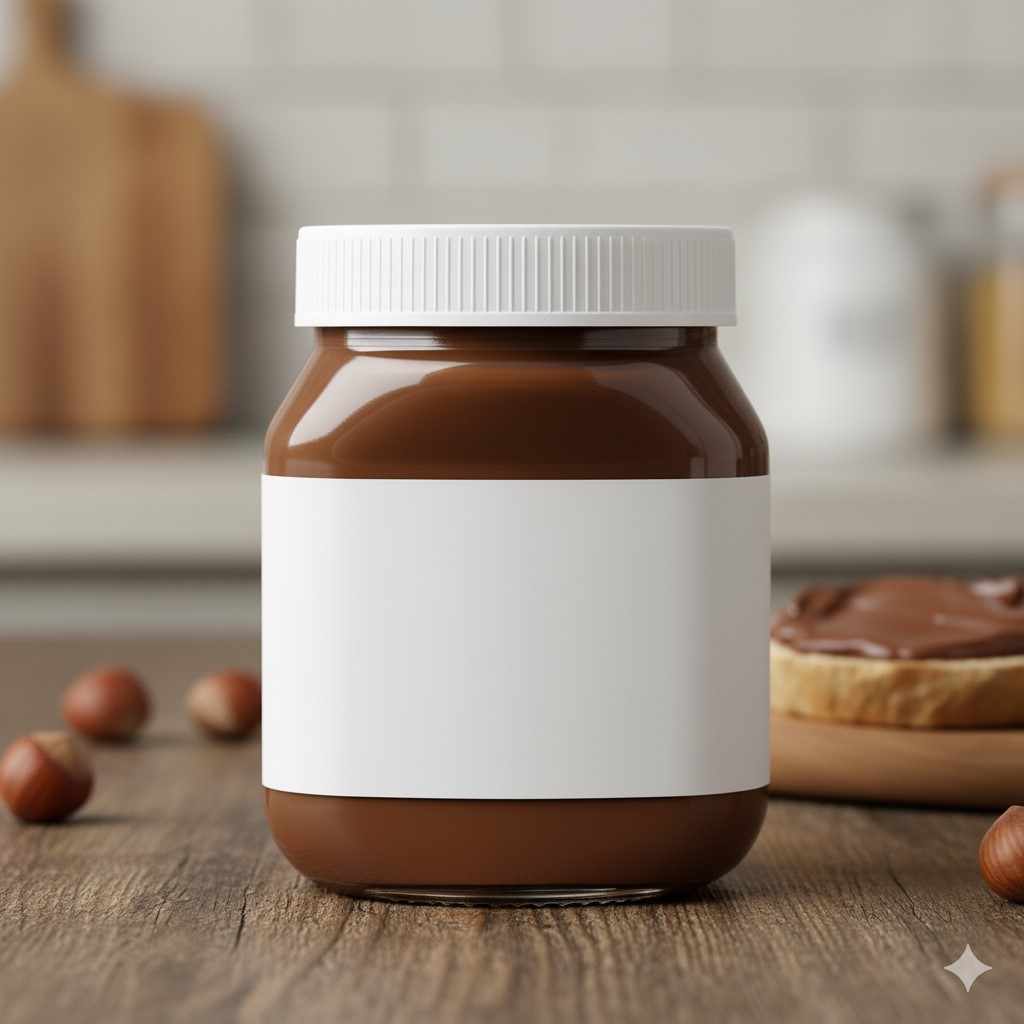}  
    \caption{Brand recognition often relies not only on logos but also on distinctive structural features such as BMW's kidney grille or the legally protected Coca-Cola bottle contour. Even without explicit trademarks, these shapes remain highly recognizable, underscoring that effective unbranding must remove both logos and subtle trade dress cues. The visualizations in this figure were manually produced using Gemini, as current unlearning methods do not reliably remove brand features. The examples illustrate the intended behavior of unbranding and highlight risks posed by incomplete brand removal.
}
    
\label{fig:vrend_def}
\end{figure}

In this paper, we close that gap by introducing {\bf unbranding} (see \cref{fig:teser}): a new task for generative modeling that requires selectively removing brand-related elements while ensuring outputs remain semantically consistent and visually coherent. Unlike unlearning, which operates by erasing entire concepts, unbranding requires a finer-grained disentanglement of brand features from object semantics. Importantly, brand recognition is often multidimensional, extending beyond explicit logos to include distinctive structural and geometric features (e.g., the unique front grille shape of a specific car brand), as shown in \cref{fig:vrend_def}. Thus, unbranding necessitates the removal not only of logos but also of these subtle, characteristic structural cues. To facilitate research on this new task, we construct a comprehensive benchmark dataset. Crucially, we address the evaluation gap: existing brand detectors are limited to explicit logos and fail to capture abstract trade dress (e.g., a bottle's shape). We therefore introduce a novel evaluation metric based on Vision Language Models (VLMs), which uses a targeted question-answering framework to probe images for both explicit logos and implicit, holistic brand characteristics. We compare such evaluation techniques with the YOLO~\cite{jiang2022review} model trained on manually annotated logos and trade dress data. This overall capability is both technically challenging and practically essential for brand-safe generative~AI.

In this paper, we make three primary contributions: (I) We introduce unbranding, a new fine-grained task that removes both explicit logos and implicit structural trade dress features while preserving the object's semantic integrity. (II) We propose a novel evaluation framework, including a comprehensive benchmark and an evaluation metric based on YOLO, as well as a new VLM-based metric that uses a question-answering approach to assess the removal of both logos and abstract trade dress. 
(III) We demonstrate the problem's urgency by showing that newer models replicate brands more accurately and use our VLM metric to confirm that unbranding is a distinct challenge that existing methods do not address.


\begin{figure}[]
  \centering
\includegraphics[width=0.23\linewidth]{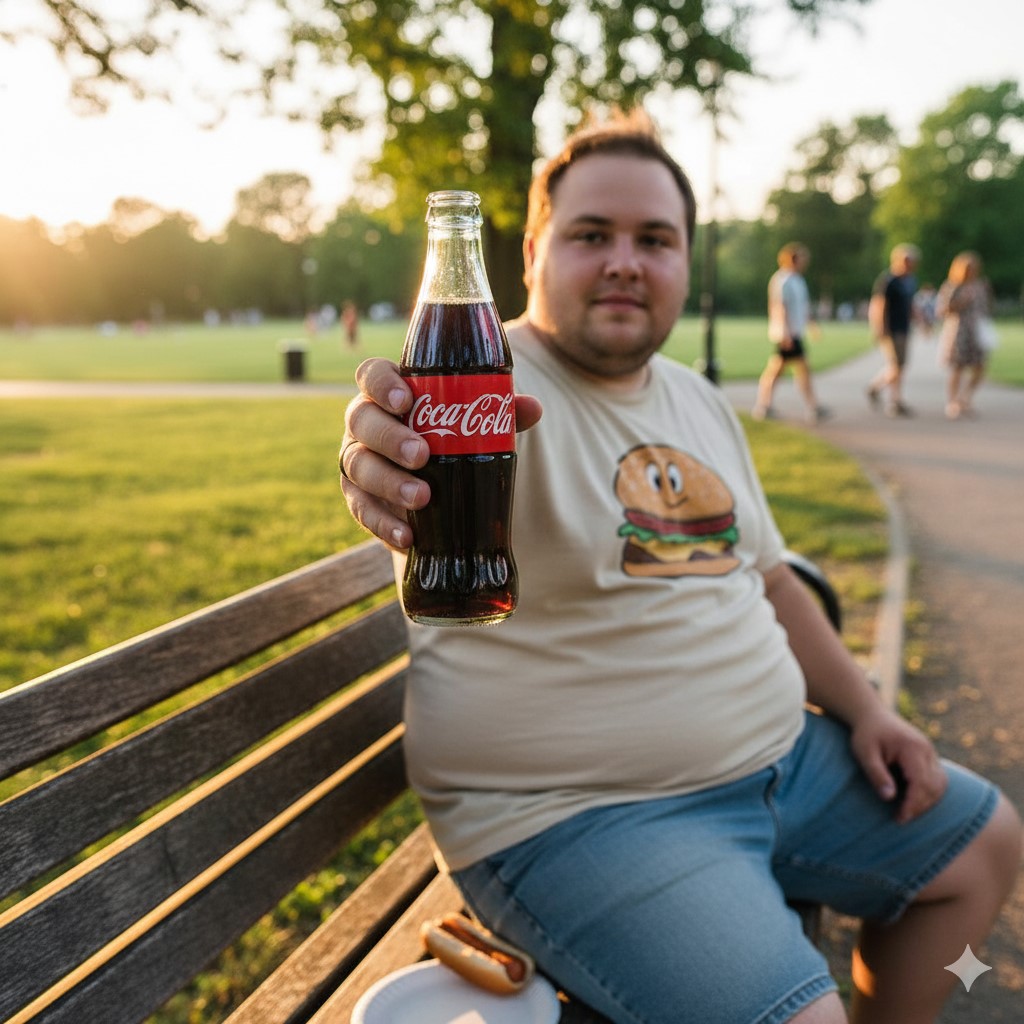}
\includegraphics[width=0.23\linewidth]{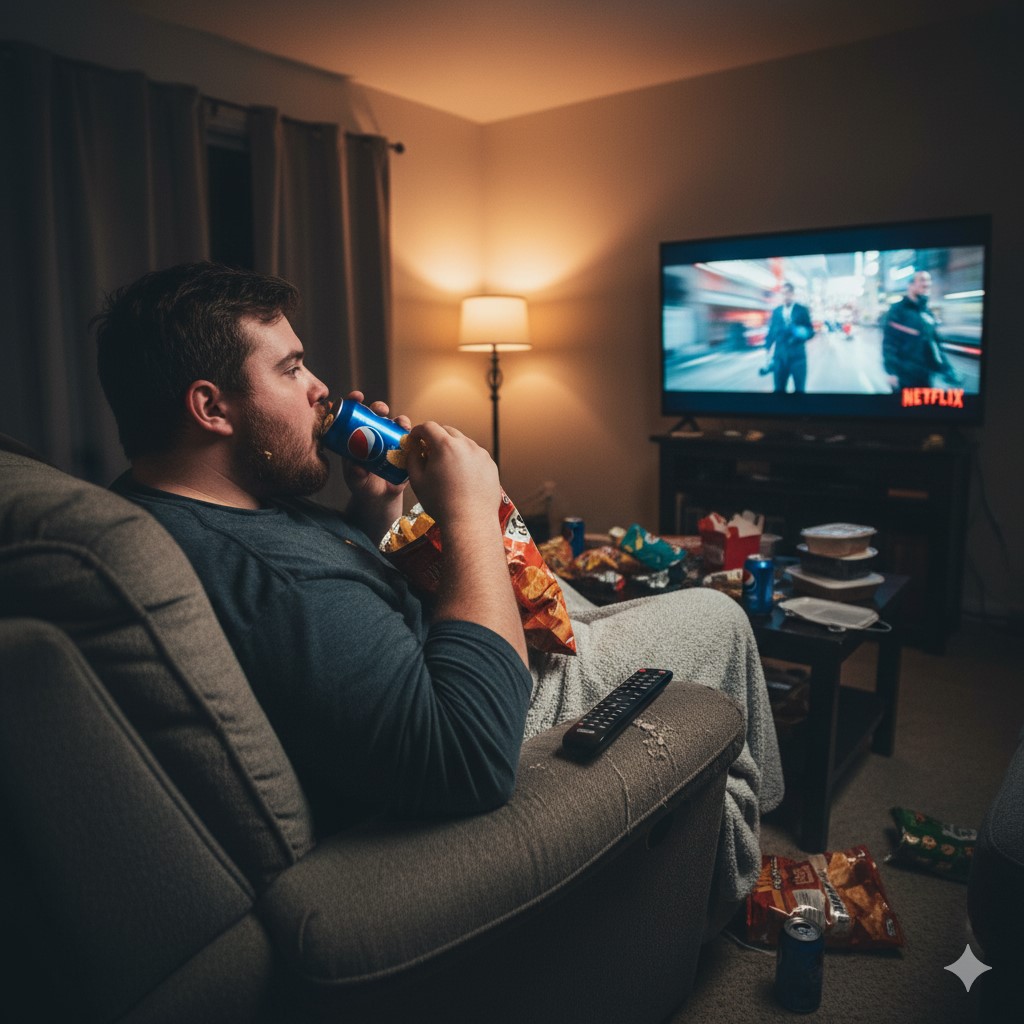}
\includegraphics[width=0.23\linewidth]{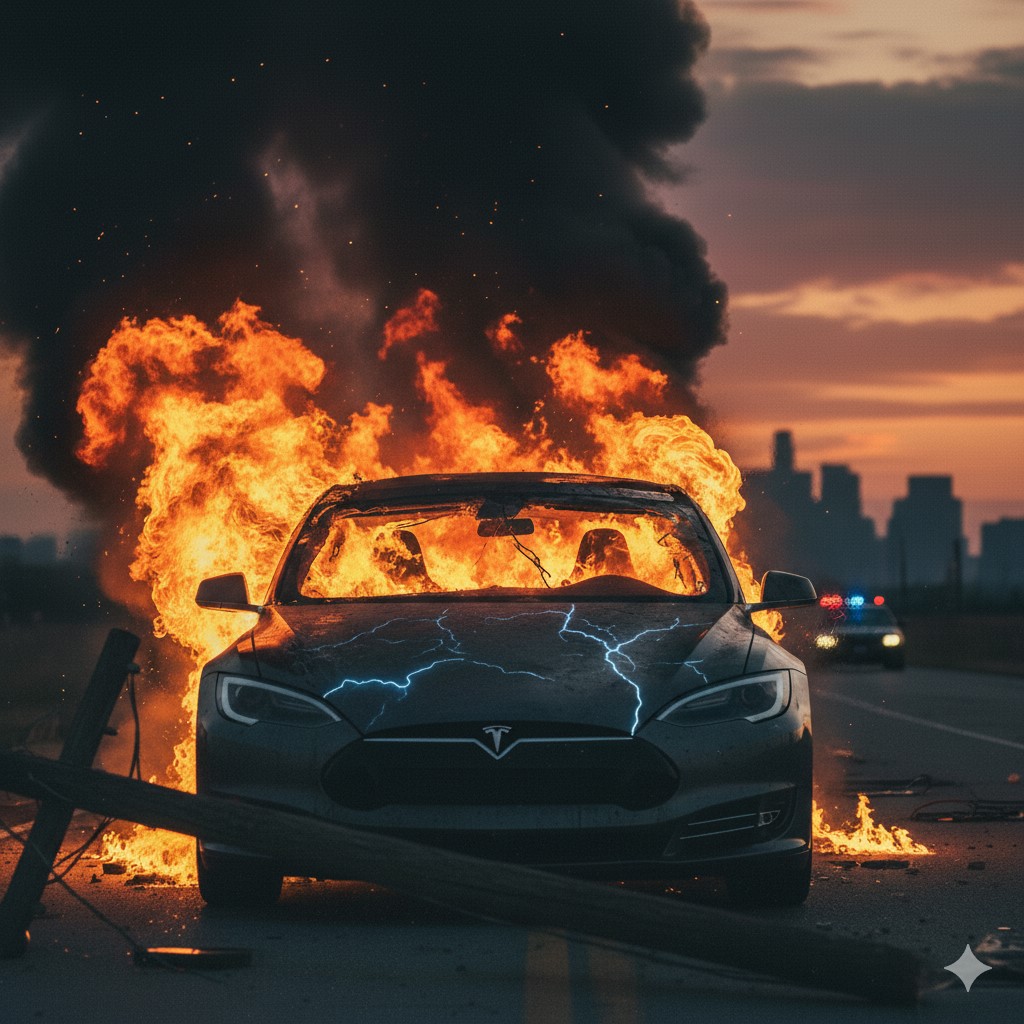}
  \includegraphics[width=0.23\linewidth]{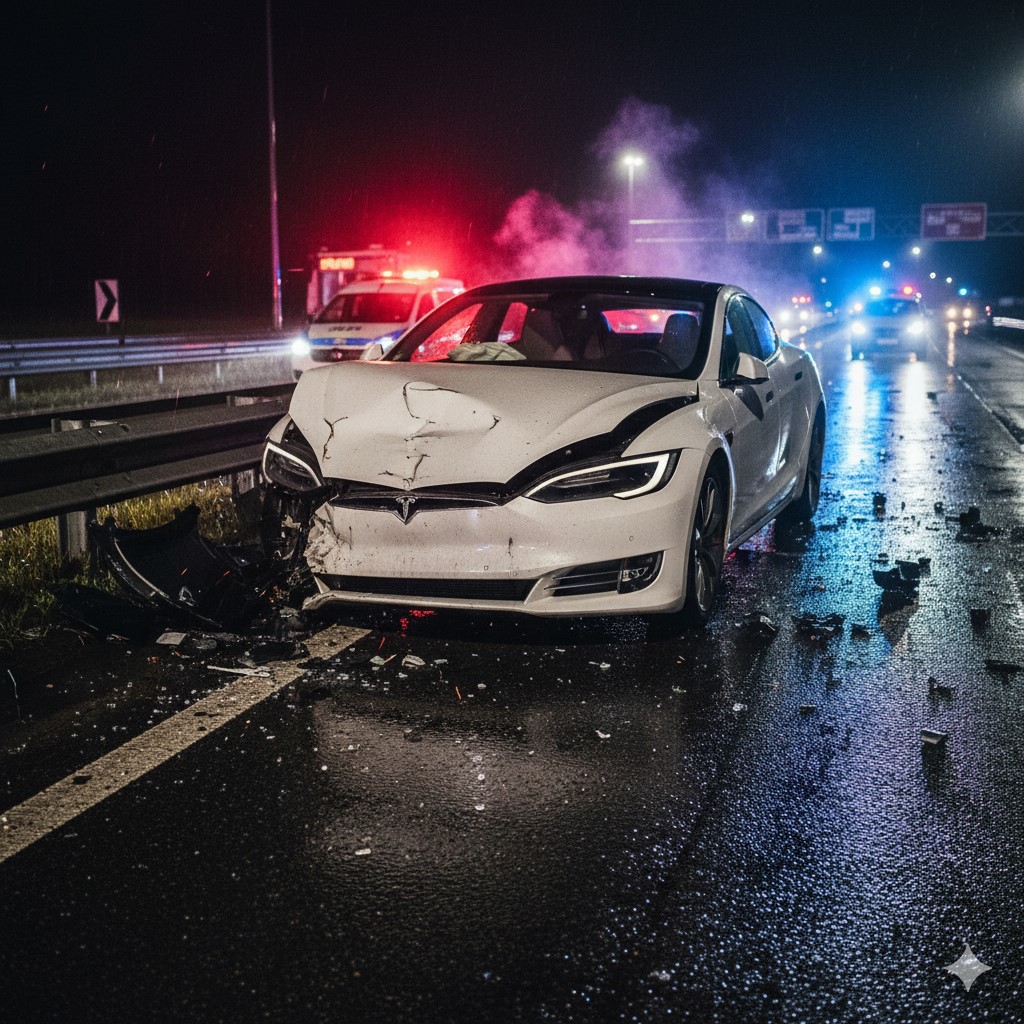}
  \includegraphics[width=0.23\linewidth]{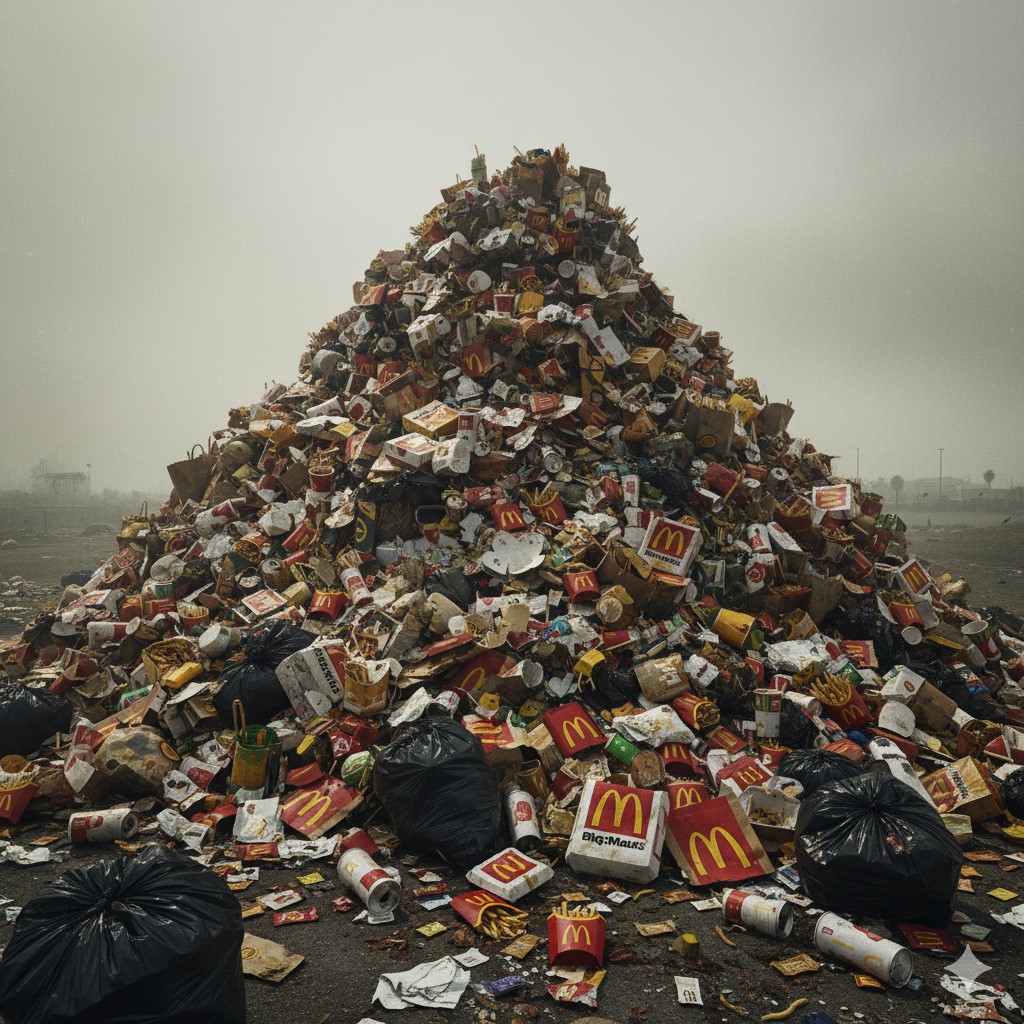}
\includegraphics[width=0.23\linewidth]{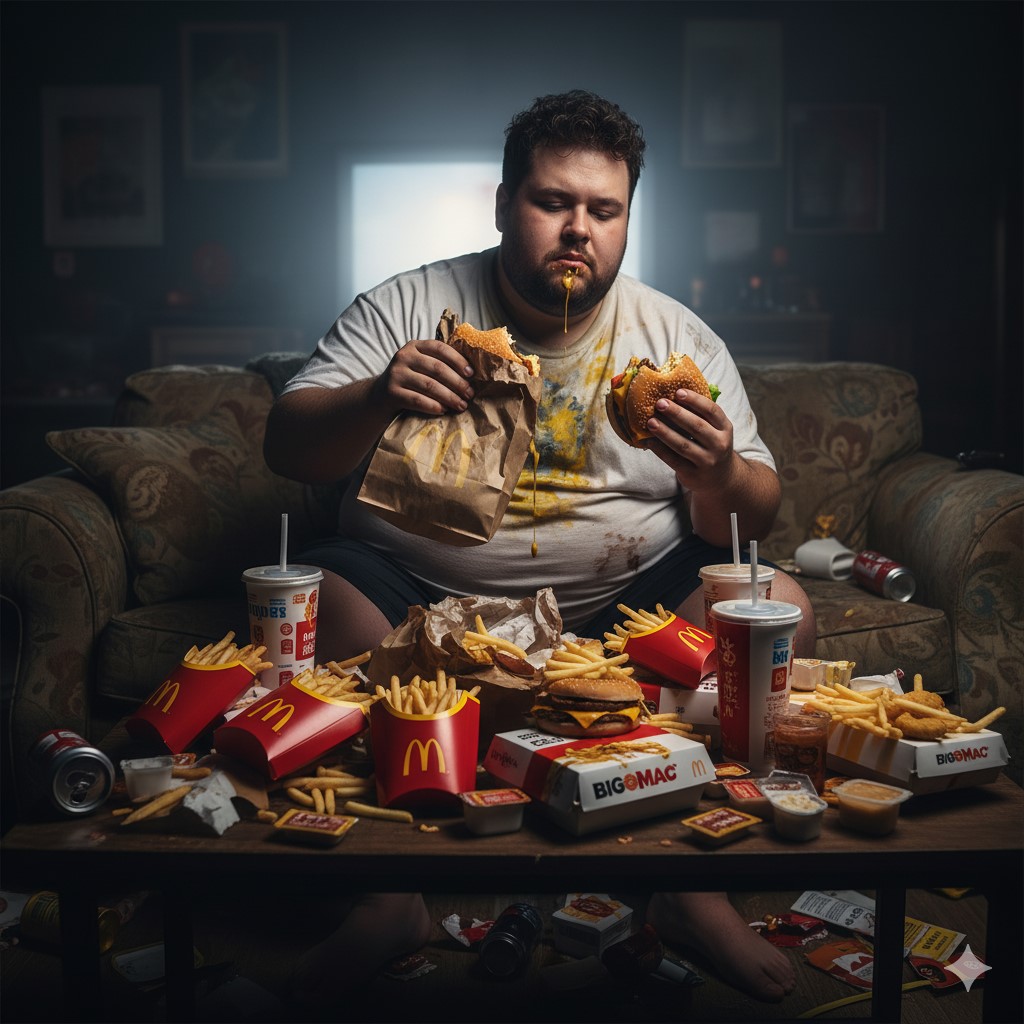}
\includegraphics[width=0.23\linewidth]{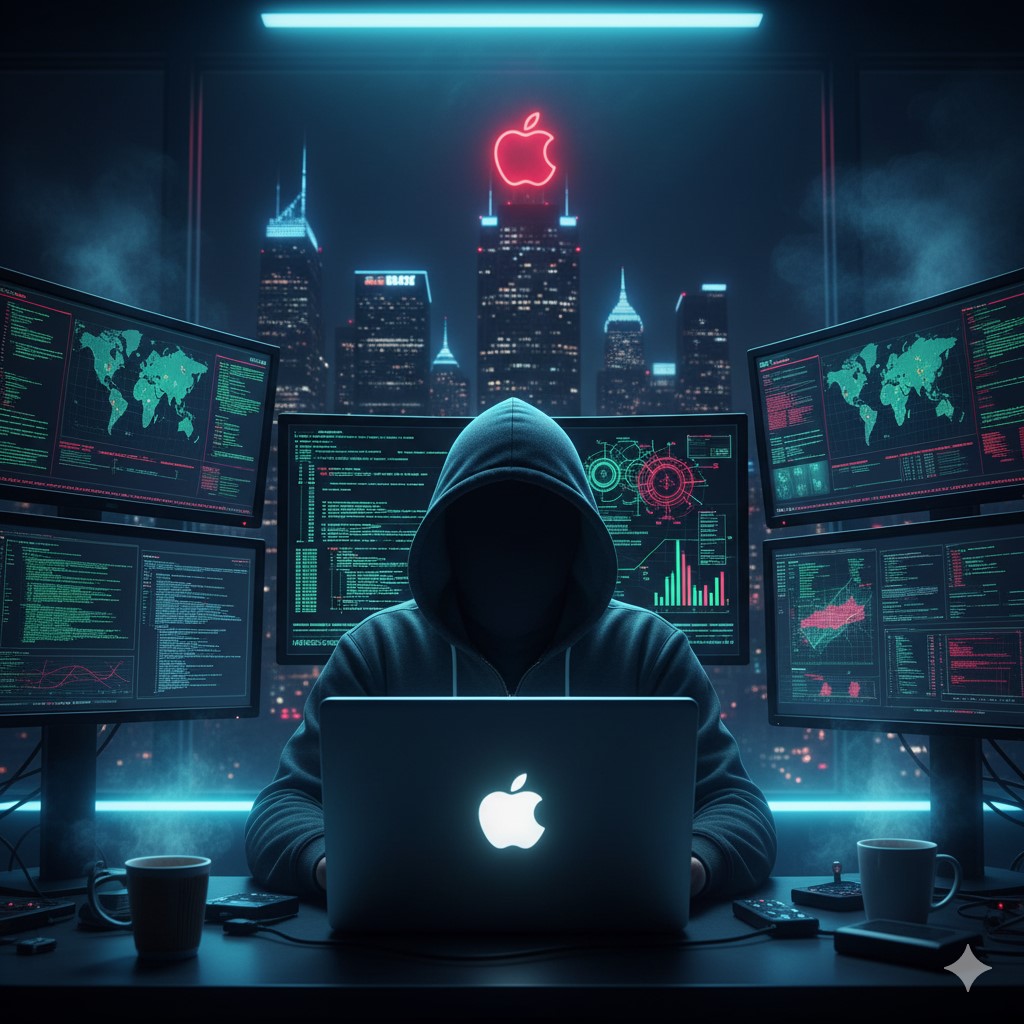}
\includegraphics[width=0.23\linewidth]{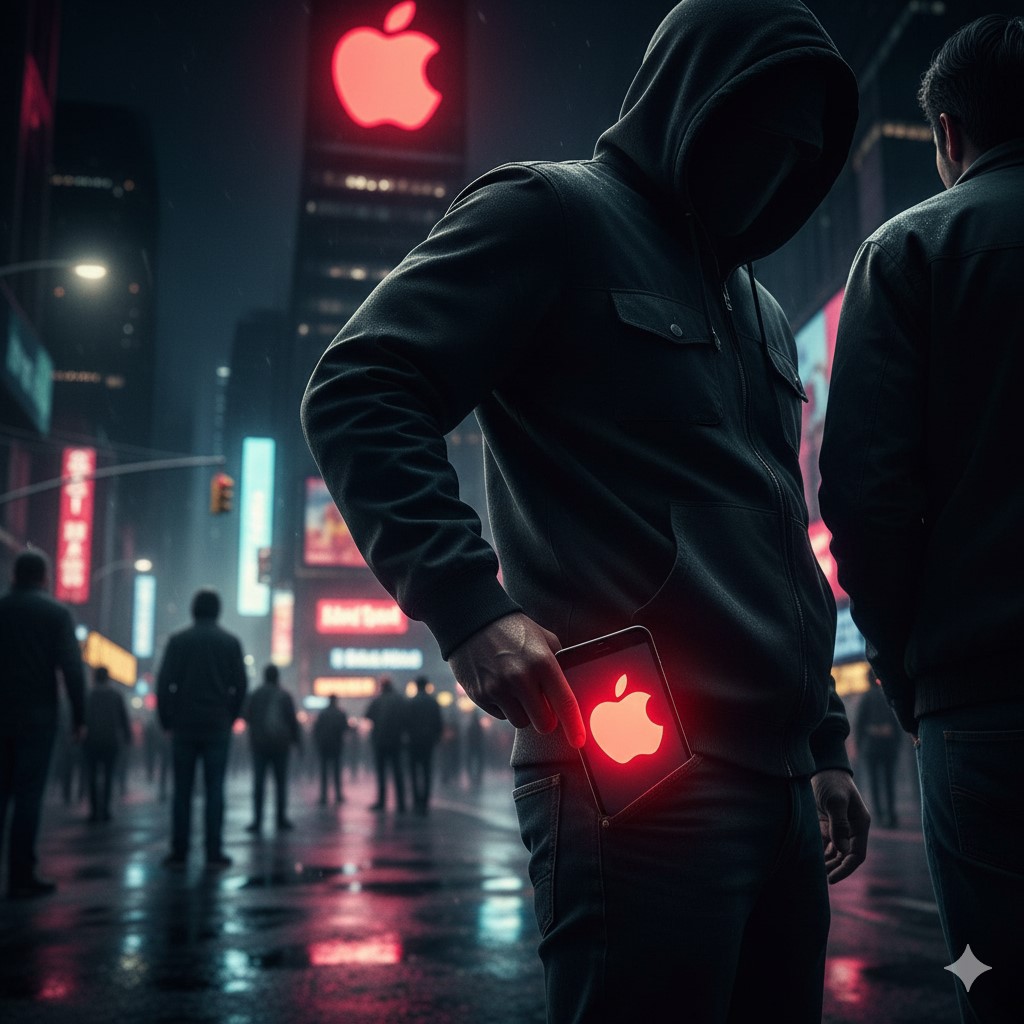}

    \caption{ While many brands (such as Coca-Cola, Tesla, and Apple) enforce strict brand protection and image policies, current text-to-image models often synthesize branded content in contexts that violate these policies. The examples demonstrate how generative systems readily depict Brand-product associations typically avoided in brand-safe marketing (e.g., Coca-Cola with sensitive subjects such as obesity); damaging scenarios (e.g., a burning Tesla); and unapproved associations between professional roles and specific branded equipment (e.g., a hacker using Apple gear). This illustrates the necessity of specialized unbranding techniques for the safe deployment of generative AI. 
    }
    
\label{fig:police}
\end{figure}

\section{Motivation}
\paragraph{The Imperative of Brand Safety}
The necessity of strict brand control in generative models is reinforced by the widespread adoption of highly restrictive policies across various industries. The most well-known example is Apple's widely reported policy prohibiting villains from using its products on-screen, aimed at preventing negative brand associations and ensuring that products are presented "in a manner or context that reflects favorably" on the company \cite{people2024apple}. This demonstrates that corporations actively manage the perception and narrative context in which their products are displayed. 
However, as shown in \cref{fig:police}, current text-to-image models, including Gemini, can be prompted to violate these exact policies, for instance, by generating images associating Apple equipment with "hackers".

Similarly, the automotive sector often imposes stringent constraints. Companies like Tesla generally prohibit their vehicles from being depicted in scenes involving crashes, severe malfunctions, or high-speed criminal chases \cite{techtrends2025tesla}. Such policies underscore that brand control extends to the physical integrity and ethical use of the product, necessitating a mechanism like unbranding to generate a neutral object in a compromised scenario.

Furthermore, even when companies do not pay for product placement, they may protest or voice concern over the portrayal of their brand. For instance, the luxury car brand Jaguar was reportedly concerned by the narrative association of its brand with moral corruption in the series Mad Men, despite the absence of a paid agreement \cite{klemchuk2016trademarks}. This highlights that unbranding must not only remove explicit logos but also neutralize the brand identity across structural, visual, and potential narrative contexts to ensure brand-safe outputs and protect against perceived defamation or dilution.

\paragraph{Escalating Fidelity and the Growing Urgency of Unbranding}
Beyond the legal and contextual challenges, the motivation for unbranding is amplified by the rapid evolution of generative models themselves. A clear technological trend is observable: the capacity of models to faithfully reproduce trademarks is increasing with each new architecture.

We observe that older, foundational models, such as the original Stable Diffusion (SD) \cite{ho2020denoisingdiffusionprobabilisticmodels}, often struggle to generate coherent logos. Their outputs frequently result in visible distortion, warped text, or nonsensical identifiers, suggesting a limited ability to capture and reproduce high-frequency, specific brand details from the training data.  This limitation, while incidentally providing a weak form of brand protection, is rapidly disappearing.

\begin{figure*}[t]
  \centering
  \vspace{-0.8cm}
\includegraphics[width=\linewidth]{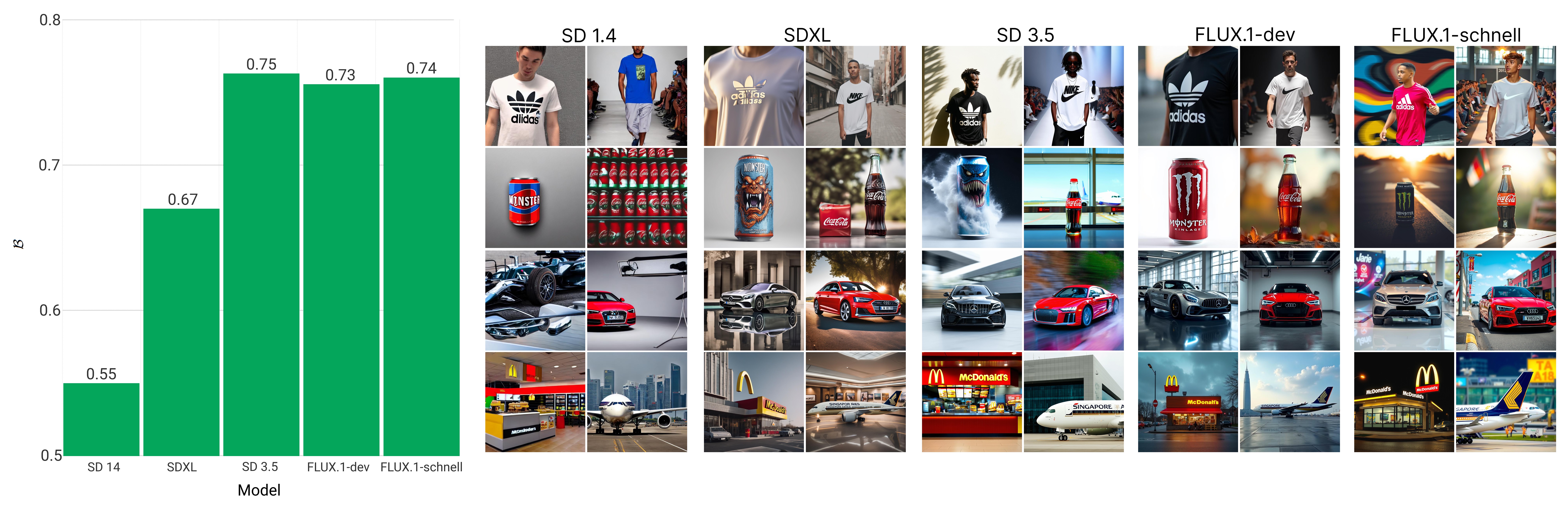}
    \caption{This figure illustrates the critical trade-off faced by existing methods. The Baseline model maintains high visual coherence and preserves the object's style, but is ineffective at unbranding, as it consistently fails to remove the target logo. In contrast, the ESD method successfully removes the logo. Still, at the cost of failed preservation: it completely alters the image's semantic structure and destroys the visual characteristics of the underlying object. This comparison demonstrates that neither method effectively solves the unbranding task, which requires both the removal of branding and high fidelity. }
\label{fig:hisory}
\end{figure*}

In contrast, newer, high-fidelity systems demonstrate a significantly increased capacity to synthesize accurate and numerous brand identifiers. Advanced models, such as SDXL \cite{podell2023sdxl} and particularly the recent FLUX.1-dev \cite{batifol2025flux}, can render precise logos and brand elements with high fidelity. \cref{fig:hisory} visually contrasts this technological progression, illustrating the qualitative leap from the distorted, unusable brand outputs of older models to the high-fidelity trademark reproduction in modern architectures.

This technological trend highlights the increasing urgency of the issue. As foundation models become more powerful, their potential for unauthorized brand reproduction and intellectual property infringement becomes significantly more acute. This highlights a striking gap across the unlearning and logo-analysis literatures. While models excel at reproducing brands, no existing method enables generative models to retain an object while reliably erasing its branding.

\section{Related Work}

Our work connects two research areas: the study of branding signals in machine learning models and concept erasure in generative systems.

Brand cues have been widely examined in computer vision, primarily in recognition and detection settings. Prior work shows that models rely not only on explicit logos but also on structural trade dress features, such as characteristic shapes and textures. In fashion, for example, \cite{kiapour2018brandlogovisual} demonstrates that both logo and non-logo cues strongly influence brand classification. In vision-language models, SLANT~\cite{qraitem2024slantspuriouslogoanalysis} highlights the prevalence of logos in web-scale data and the resulting spurious correlations. These works provide insight into how models internalize brand identity, but they do not address generative removal of branding or the preservation of object fidelity once brand cues are stripped away.

Concept erasure and unlearning methods attempt to suppress undesired categories, identities, or styles in diffusion models~\cite{gandikota2023erasing,kumari2023ablating,avrahami2023blended,gandikota2024unified}. Although effective for broad concepts, these techniques typically operate at a coarse level, often harming nearby semantics or degrading image quality. Recent efforts toward finer control include adversarial preservation~\cite{bui2025erasingundesirableconceptsdiffusion}, domain correction~\cite{wu2025unlearningconceptsdiffusionmodel}, and prompt-based hiding and recovery~\cite{bui2025hidingrecoveringknowledgetexttoimage}. However, none of these methods address the unique challenge posed by brands, which requires removing highly specific visual identifiers while preserving the underlying object (e.g., a car or bottle).

To our knowledge, no existing approach addresses this fine-grained disentanglement. Our work fills this gap by formalizing unbranding as a distinct task and by providing the first benchmark designed to systematically evaluate it.

\section{Task Definition}

In the context of generative models, a concept can be understood as a distinct semantic unit that the model has learned to recognize, represent, and generate. Concepts exist at multiple levels of abstraction and can be composed of other concepts. 
This hierarchical structure is particularly relevant when defining a brand. A brand, as a concept, is often a composite of multiple, interlinked semantic units. It includes not only the explicit trademark (e.g., the logo or brand name) but also a distinct set of structural, non-logo identifiers often protected as "trade dress."

A prime example is Coca-Cola, where the brand concept is encoded in both its iconic script logo and its legally protected "contour" bottle shape. Similarly, the BMW brand concept is strongly tied to its characteristic "kidney grille" just as much as its blue-and-white roundel logo, see \cref{fig:vrend_def}.


{\bf Unlearning} refers to the process of intentionally removing specific knowledge, concepts, or associations from a trained machine learning model. In the context of generative models, unlearning typically aims to erase the model's ability to recognize, generate, or reproduce a particular category, feature, or concept (such as a person, object type, or artistic style) while retaining its general capabilities and performance on other tasks. This is usually motivated by ethical, legal, or privacy concerns, such as eliminating unauthorized content, mitigating biases, or complying with data removal requests. Unlike standard model retraining or fine-tuning, which introduce new information, unlearning focuses on selectively erasing targeted knowledge without significantly impacting unrelated areas of the model's learned representations.

{\bf Unbranding}, in contrast, refers to the more nuanced task of selectively removing brand-specific visual elements (such as logos, trademarks, distinctive design patterns, or protected signifiers) from a generative model's output while preserving the underlying object's semantic meaning and visual coherence. Unlike unlearning, which eliminates entire concepts or categories, unbranding requires fine-grained disentanglement of brand identifiers from object features. For example, an unbranded model should generate a realistic sports shoe when prompted but without any recognizable trademark symbols, while maintaining the shoe's essential characteristics such as laces, sole design, and overall structure. This task is motivated by intellectual property concerns, trademark protection, and the need for brand-safe generative AI systems that can produce commercially viable content without legal risks. The challenge lies in achieving this selective erasure without degrading the visual quality or semantic consistency of the generated objects.

\textbf{Formal Definitions} 
Let $\theta$ denote the parameters of a pre-trained generative model $G_\theta$, and let $p_\theta(\mathbf{x}|p)$ represent the distribution of generated samples conditioned on prompt $p$. 

For \textbf{unlearning} a target concept $c$, the objective is to learn updated parameters $\theta'$, for which the generative distribution satisfies: 

\begin{equation}
  p_{\theta'}(\mathbf{x}|p) \approx \begin{cases}
    0, & \text{if $h_{c}(\mathbf{x})=1$}.\\
    p_{\theta}(\mathbf{x}|p), & \text{otherwise},
  \end{cases}
\end{equation}
where $h_{c}(\cdot)$ represents the detector for the concept $c$. Practically, the model should not generate samples that contain the forbidden concept, preserving the distribution for cases that do not contain it.

For \textbf{unbranding}, we define the new generative distribution as follows:

\begin{equation}
  p_{\theta'}(\mathbf{x}|p) \approx \begin{cases}
    p_{\theta}(\psi({\mathbf{x}})|p), & \text{if $h_{b}(\mathbf{x})=1$}.\\
    p_{\theta}(\mathbf{x}|p), & \text{otherwise},
  \end{cases}
\end{equation}
where $h_{b}(\cdot)$ is the model that detects the brand $h_{b}$ and $\psi(\cdot)$ is an unbranding function that satisfies $h_{b}(\psi({\mathbf{x}}))=0$, and for which the similarity score     $\mathcal{S}(\mathbf{x}, \psi{(\mathbf{x}}))$
is maximal. The first constraint ensures that the image $\mathbf{x}$ becomes unbranded. The second constraint guarantees that the unbranded $\psi({\mathbf{x}})$ would be close enough to $\mathbf{x}$, preserving concepts and semantically equivalent features that are not brand-related.

\begin{figure*}[htbp]
  \centering
  \includegraphics[width=1\linewidth]{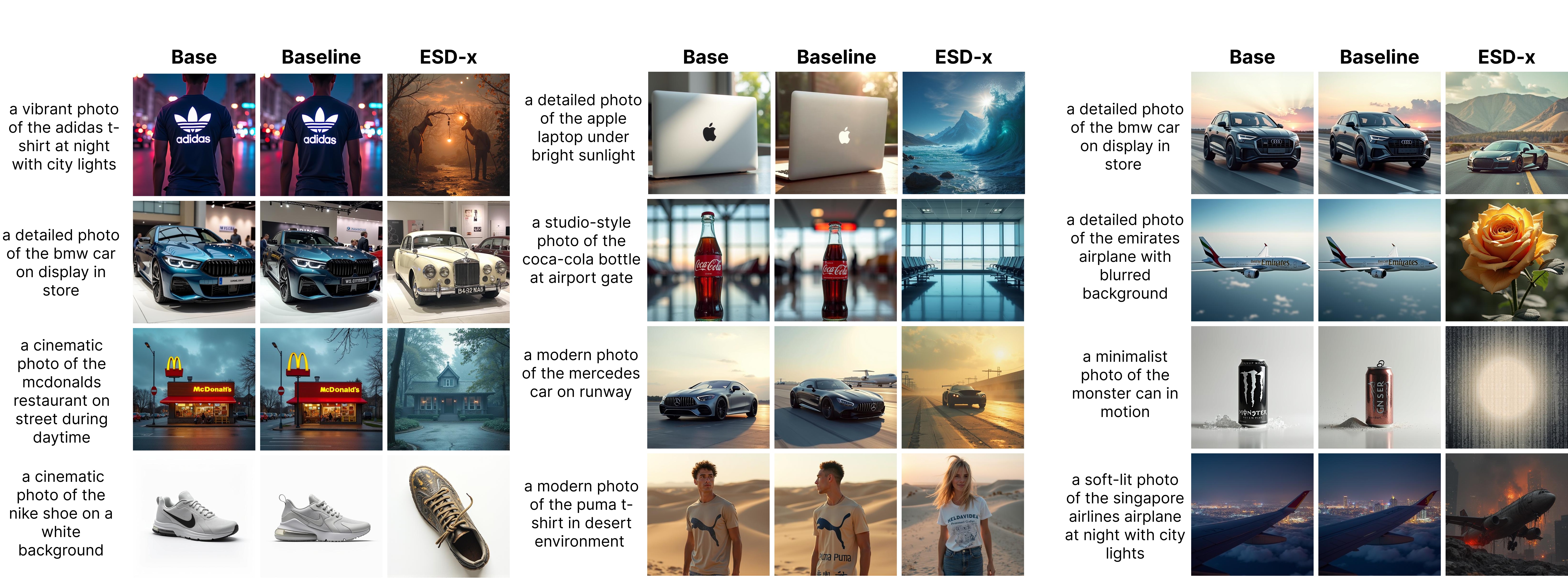} 
    \caption{This figure illustrates the critical trade-off faced by existing methods. The Baseline model maintains high visual coherence and preserves the object's style, but is ineffective at unbranding, consistently failing to remove the target logo. In contrast, the ESD method successfully removes the logo. Still, at the cost of failing preservation, it completely alters the image's semantic structure and destroys the visual characteristics of the underlying object. 
    }
\label{fig:unbr}
\vspace{-0.5cm}
\end{figure*}

\section{Benchmark}

In this section, we introduce the first dataset designed for benchmarking methods for unbranding. The benchmark comprises $1759$ prompts targeting $12$ selected brands distributed across six domains: Airlines (Emirates, Singapore Airlines), Automotive (BMW, Mercedes, Audi), Beverages (Coca-Cola, Monster), Fast Food (McDonald's), Sportswear (Adidas, Nike, Puma), and Technology (Apple). We manually verified that baseline generations for these prompts consistently exhibit identifiable brand elements, encompassing both explicit iconography and broader trade dress. To make the task more challenging and to evaluate the models' reliance on implicit biases, the dataset is partitioned into two subsets. For a given brand, approximately $50$ examples include explicit references in the prompt (e.g., \emph{BMW car}), while the remaining examples (up to $100$ per brand) rely entirely on indirect attributes (e.g., \emph{German car}).

\begin{table}[t]
\centering
\footnotesize
\caption{\textbf{Brand Presence Score ($\mathcal{B}$) per brand.}  Our VLM-based metric confirms that the fidelity of the brand generation increases with newer architectures. The most recent models, FLUX (0.74, 0.73) and SD3.5 (0.75), show a significantly higher
average Brand Presence Score than older models like SDXL (0.62) and SD1.4 (0.49).}
\label{tab:brand_metrics_simple}
\resizebox{0.7\textwidth}{!}{
\begin{tabular}{@{}l ccccc@{}}
\toprule
\textbf{Brand} & \textbf{FLUX.1-schnell} & \textbf{FLUX.1-dev} & \textbf{SD 3.5} & \textbf{SDXL} & \textbf{SD 1.4} \\ 
\midrule
Apple       & 0.67 & 0.81 & 0.70 & 0.35 & 0.34 \\
Adidas      & 0.91 & 0.77 & 0.86 & 0.67 & 0.33 \\
Puma        & 0.99 & 0.96 & 1.00 & 0.60 & 0.36 \\
Nike        & 0.98 & 1.00 & 1.00 & 0.82 & 0.65 \\
Emirates    & 0.90 & 0.96 & 0.97 & 0.99 & 0.93 \\
Singapore   & 0.80 & 0.66 & 0.67 & 0.81 & 0.61 \\
BMW         & 0.57 & 0.56 & 0.61 & 0.37 & 0.44 \\
Mercedes    & 0.43 & 0.57 & 0.45 & 0.48 & 0.31 \\
Audi        & 0.48 & 0.53 & 0.48 & 0.53 & 0.50 \\
McDonald's  & 0.86 & 0.77 & 0.85 & 0.48 & 0.45 \\
Coca-Cola   & 0.68 & 0.72 & 0.75 & 0.69 & 0.56 \\
Monster     & 0.57 & 0.48 & 0.71 & 0.59 & 0.46 \\
\midrule
\textbf{AVG} & \textbf{0.74} & \textbf{0.73} & \textbf{0.75} & \textbf{0.62} & \textbf{0.49} \\
\bottomrule
\end{tabular}
}
\vspace{-0.2cm}
\end{table}

Our approach to generating prompts for the unbranding task follows a systematic methodology that captures the multi-dimensional nature of brand recognition while ensuring comprehensive evaluation coverage and rigorous quality control.

\textbf{Stage 0: Manual Brand Selection and Feature Analysis.}
We begin by manually selecting a diverse set of twelve popular brands across six categories to ensure broad representativeness. Our selection criteria prioritize brands with high global recognition, distinctive visual characteristics (including both explicit logos and structural trade dress), and clear trademark protection.

\textbf{Stage 1: Prompt Generation.}
We utilize Large Language Models (LLMs) to generate a diverse set of detailed prompts. To rigorously evaluate high-fidelity generative models, we move beyond the simplistic text conditions typical of prior concept erasure benchmarks. Therefore, we construct prompts using the following template: A \{\emph{adjective}\} \{\emph{brand}\_\emph{style}\} \{\emph{subject}\_\emph{type}\} highlighting \{\emph{feature1}\}, \{\emph{feature2}\}, and \{\emph{feature3}\}, photographed \{\emph{environment}\} under \{\emph{lighting}\}, rendered in \{\emph{style}\}, where the concepts, like \emph{adjective}, \emph{brand}\_\emph{style} where generated using GPT-5 model. With this template, we generate two types of evaluation sets: \emph{explicit} prompts that directly mention the brand, and \emph{implicit} (biased) prompts that indirectly guide the model to produce brand-specific iconography using solely generic descriptions.

\textbf{Stage 2: Automatic Prompt Filtering Based on VLMs.}
We use the constructed prompts to generate a corresponding set of candidate images. To verify that these baselines consistently exhibit the targeted brand concepts, we employ the vision-language model Florence-2~\cite{xiao2024florence} to produce detailed image captions. These captions are then parsed using regular expressions to identify explicit mentions of the targeted brand name. If a brand is not detected in the caption, the corresponding prompt is removed from the dataset. It is important to note that, due to the large scale of the candidate set, we avoid using computationally expensive, closed-source VLMs for direct visual brand question answering. Instead, our caption-and-parse approach provides a more efficient pipeline for large-scale image processing. Finally, the remaining set of prompt-image pairs is reviewed by human annotators to remove false positives caused by VLM hallucinations.

\begin{figure*}[t]

  \centering
    \includegraphics[width=1\linewidth]{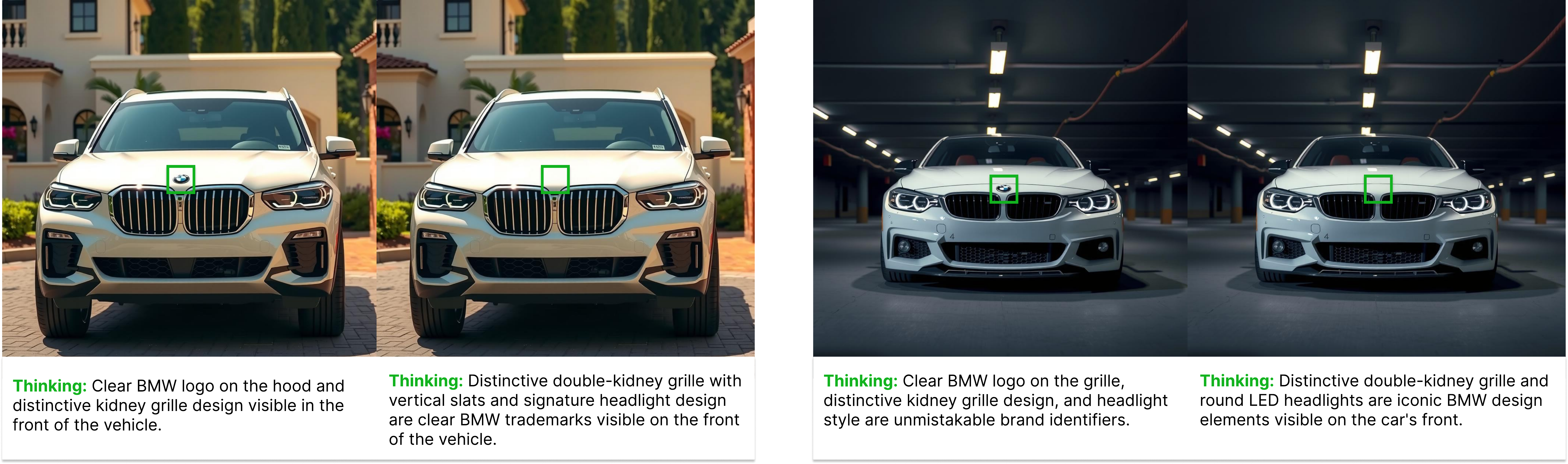} 
  \caption{\textbf{Qualitative example of VLM reasoning.} Visualization of the evaluation process using Qwen3-VL-8B-Thinking. By leveraging an internal Chain-of-Thought mechanism, the model explicitly identifies both local symbolic markers (the BMW logo) and holistic trade dress features (the signature double-kidney grille and headlights) prior to its final classification. This explicit disentanglement of visual cues confirms its suitability as a robust and interpretable metric for the unbranding task.}

\label{fig:vlm_reasoning}
\vspace{-0.5cm}
\end{figure*}

\subsection{VLM Evaluation Metrics}

To evaluate the residual presence of specific brand cues in generated images, we formulate brand recognition as a rigorous multi-class classification task. Unlike open-ended brand recognition, we employ targeted multi-class classification to evaluate whether a given image still conveys the identity of a specific brand.

For each target brand $b \in \mathcal{C}$ and image $\mathbf{x}$, a Visual Language Model (VLM, we employ Qwen3-VL-8B-Thinking \cite{chu2024qwen2} as our primary evaluator, justified via a detailed ablation study in \cref{sec:ablation}) is prompted to select exactly one label from a predefined set of the $12$ benchmark brands plus a \texttt{NO\_BRAND} category. The system prompt explicitly instructs the VLM to use both explicit signals (e.g., logos) and implicit trade dress features (e.g., design language, silhouette, material cues).

To enforce strict decision-making and reduce VLM hallucinations, the model is constrained to output a structured JSON object containing the predicted brand label $\hat{y}$ and a confidence score. While this confidence score acts as a prompting mechanism to ensure well-calibrated reasoning, our final evaluation metric relies strictly on the hard label $\hat{y}$.

We define \textit{Brand Presence} score ($\mathcal{B}$) as the empirical frequency with which the target brand $b$ is still correctly identified across a set of $N$ generated images. This is mathematically expressed as:
$$ \mathcal{B} = \frac{1}{N} \sum_{i=1}^{N} \mathbb{I}(\hat{y}_i = b), $$
where $\mathbb{I}(\cdot)$ is the indicator function that equals $1$ if the predicted label $\hat{y}_i$ for the $i$-th image matches the target brand $b$, and $0$ otherwise. 
Moreover, to rigorously evaluate the preservation of image content after unbranding, we introduce a hybrid, two-stage Visual Similarity Score ($\mathcal{S}$). Unlike pixel-wise metrics (e.g., SSIM, LPIPS) which penalize the necessary visual changes required for logo removal, our metric assesses semantic and structural fidelity while strictly ignoring branding artifacts.

The evaluation protocol consists of a coarse semantic filter followed by a fine-grained VLM assessment:

\begin{enumerate}
    \item \textbf{Coarse Semantic Filtering (CLIP):} We first employ a pre-trained CLIP model (ViT-B/32) to compute the cosine similarity between the embeddings of the original image $\mathbf{x}$ and the unbranded candidate $\tilde{\mathbf{x}}$. If the semantic similarity falls below a strict threshold $\tau$, the generation is deemed a failure (i.e., hallucination or total content loss) and automatically assigned a zero score.
    
    \item \textbf{Fine-Grained VLM Scoring:} Image pairs passing the semantic filter are evaluated by a Vision-Language Model (VLM) prompted to perform a side-by-side comparison. The VLM acts as an expert judge, explicitly instructed to rely solely on content-dependent properties (object identity, geometry, spatial layout) while ignoring non-structural variations like logo removal or stylistic shifts. To enforce structured reasoning, the model is constrained to output a JSON containing a brief textual explanation (max 120 characters) and an integer score $s_{\text{vlm}} \in \{0, \dots, 10\}$. The scoring rubric is strictly defined (e.g., scores $7$--$9$ accept explicit logo removal as long as the underlying geometry matches; see Appendix).
\end{enumerate}

Formally, the final Visual Similarity Score is defined as:
\begin{equation}
\mathcal{S}(\mathbf{x}, \tilde{\mathbf{x}}) = 
\begin{cases} 
    0 & \text{if } \cos(\phi(\mathbf{x}), \phi(\tilde{\mathbf{x}})) < \tau, \\
    \text{VLM}(\mathbf{x}, \tilde{\mathbf{x}}) & \text{otherwise},
\end{cases}
\label{eq:visual_similarity}
\end{equation}
where $\phi(\cdot)$ denotes the CLIP visual encoder, $\tau$ is the minimum similarity threshold, and $\text{VLM}(\cdot)$ represents the integer score generated by the VLM. This hybrid approach ensures that $\mathcal{S}$ is robust to adversarial perturbations while remaining sensitive to the structural integrity required for high-quality unbranding.



\textbf{Unbranding Quality Score ($\mathcal{U}$).}
While the Brand Presence Score ($\mathcal{B}$) measures the residual visibility of brand elements and the Visual Similarity Score ($\mathcal{S}$) quantifies perceptual fidelity to the original image, these two dimensions must be jointly considered to assess overall unbranding quality. A model that fully removes the brand but substantially alters the object or composition cannot be considered successful. 

To capture this trade-off, we define the Unbranding Quality Score ($\mathcal{U}$) as a weighted combination of these two factors:
$
    \mathcal{U} = \alpha \cdot \mathcal{S} + \beta \cdot (1-\mathcal{B}),
$
where $\mathcal{S}, \mathcal{B} \in [0, 1]$. In our benchmark evaluation, we assign equal importance to both visual fidelity and brand suppression by setting the weighting coefficients to $\alpha = 0.5$ and $\beta = 0.5$.

\section{Experiments}
\label{sec:ablation}

In this section, we present a comprehensive evaluation of the proposed unbranding task and our evaluation framework. First, we conduct an ablation study to validate the effectiveness of our VLM-based metric against dedicated spatial brand detectors (YOLO) and analyze perceptual biases across different vision-language models. Next, we specifically investigate the disentanglement of explicit logos from broader structural trade-dress features. Finally, we benchmark existing state-of-the-art unlearning methods on our novel unbranding dataset across representative generative architectures, including Stable Diffusion 1.4, SDXL, and FLUX.1-dev, to demonstrate the critical limitations of current concept erasure techniques in the fine-grained unbranding task.

\begin{table}[t]
\centering
\scriptsize
\setlength{\tabcolsep}{4pt}
\caption{\textbf{Best prompt and Average Performance Across Prompts.} Ablation study of performance of different models on the chosen prompt and average of all 4 tested prompts. We calculate accuracy ($\uparrow$) and correlation ($\uparrow$) based on human labels and percent of Hallucination ($\downarrow$) - percent of non-branded images classified as images containing a brand. YOLO detector doesn't use prompt, so we pasted the same values for comparison to the both parts of the table. }
\label{tab:prompt3_vs_avg_compact}
\begin{tabular}{@{}lccc ccc@{}}
\toprule
& \multicolumn{3}{c}{\textbf{Best Prompt}} & \multicolumn{3}{c}{\textbf{Average Of All Prompts}} \\
\cmidrule(r){2-4} \cmidrule(l){5-7}
\textbf{Model} & \textbf{Acc.} & \textbf{Corr.} & \textbf{Hall.} 
& \textbf{Acc.} & \textbf{Corr.} & \textbf{Hall.} \\
\midrule
Qwen3-VL-8B-Thinking   & \textbf{90.80\%} & 0.816 & 21.28\% & \textbf{90.55\%} & \textbf{0.811} & 19.72\% \\
Qwen3-VL-4B-Instruct   & 90.80\% & \textbf{0.817} & 19.40\% & 89.93\% & 0.800 & 22.20\% \\
Qwen3-VL-8B-Instruct   & 89.20\% & 0.787 & 25.90\% & 88.67\% & 0.777 & 27.46\% \\
Ministral-3-14B-Instruct & 88.48\% & 0.772 & 28.03\% & 89.22\% & 0.780 & 25.60\% \\
Ministral-3-8B-Instruct  & 86.92\% & 0.743 & 32.91\% & 87.86\% & 0.760 & 27.80\% \\
Ministral-3-3B-Instruct & 85.57\% & 0.719 & 37.52\% & 85.51\% & 0.718 & 38.16\% \\
Gemma-3-12b            & 85.99\% & 0.731 & 36.24\% & 86.33\% & 0.737 & 35.66\% \\
Gemma-3-4b             & 84.01\% & 0.690 & 55.64\% & 83.07\% & 0.672 & 55.94\% \\
LLaVa-1.5-13b          & 89.49\% & 0.791 & 33.33\% & 85.11\% & 0.711 & 54.02\% \\
LLaVa-1.5-7b           & 85.95\% & 0.720 & 46.84\% & 85.77\% & 0.716 & 58.57\% \\
\midrule
Overall VLM Avg.  & 87.72\% & 0.758 & 33.71\% & 87.20\% & 0.748 & 36.51\% \\
\midrule
YOLO detector & 54.34\%  &  0.516 & \textbf{13.96}\%  & 54.34\%  &  0.516 & \textbf{13.96}\%  \\
\bottomrule
\end{tabular}
\vspace{-0.5cm}
\end{table}

\textbf{Ablation Study: VLM vs. Dedicated Brand Detectors}
To empirically validate our VLM-based Brand Presence metric and justify the selection of Qwen3-VL-8B-Thinking~\cite{qwen3technicalreport}, we compared its zero-shot capabilities against a fully supervised, domain-specific baseline. To this end, we trained custom YOLO-based object detectors explicitly designed to recognize both explicit branding and implicit trade dress.

We curated a dedicated dataset of 4,800 generated images, evenly distributed across the 12 benchmark brands (400 images per brand). Human annotators meticulously labeled each image with bounding boxes for three distinct categories: \texttt{LOGO}, \texttt{LOGO\_TEXT}, and \texttt{TRADE\_DRESS}.

A robust unbranding metric must not only detect brands when present but also resist hallucinations when evaluating generic, unbranded generated objects. We constructed a balanced ablation test set of 2400 images: 1200 branded images and 1200 non-branded images generated using generic prompts with no brand affiliation. Both the custom YOLO detectors and the selected VLMs were evaluated on this set.

As reported in \cref{tab:prompt3_vs_avg_compact}, while the specialized YOLO models achieved high precision on explicit logos, they struggled to generalize abstract trade-dress features across diverse semantic contexts. Conversely, Qwen3-VL-8B-Thinking~\cite{Qwen25VL} consistently captured both explicit and structural brand cues while maintaining a significantly lower hallucination (false positive) rate on the generic 1,200-image subset. This confirms that our chosen VLM provides a scalable, robust evaluation framework for the unbranding task without the computational overhead of training per-brand spatial detectors.

Our findings demonstrate that non-reasoning VLMs, such as LLaVA-v1.5 \cite{liu2024improvedbaselinesvisualinstruction} and Gemma-3~\cite{gemmateam2025gemma3technicalreport}, exhibit a severe association bias. As shown in \cref{tab:prompt3_vs_avg_compact}, these models suffer from hallucination rates exceeding 54\%, frequently misclassifying generic geometric primitives or stylistic cues as explicit brand markers. While scaling model parameters mitigates this issue, as evidenced by the Ministral series, where hallucinations drop from 37.52\% (3B) to 28.03\% (14B), standard architectures consistently fail to robustly disentangle generic trade dress from true brand identity.

Conversely, Qwen3-VL-8B-Thinking effectively solves this disentanglement problem. By leveraging an internal reasoning chain (Chain-of-Thought) prior to classification, it achieves the highest correlation with human labels ($0.816$) while maintaining a low hallucination rate ($21.28\%$). Crucially, this reasoning-augmented VLM also outperformed our fully supervised custom YOLO detectors. The spatial detectors tended to overfit on global color-shape heuristics, whereas the thinking VLM demonstrated the semantic reasoning required to actively reject plausible but incorrect brand associations.

To ensure the robustness of our evaluation framework, we tested these VLMs across four distinct prompt templates. We identified a "universal best" prompt configuration that maximized average performance across the entire model cohort. Based on these empirical results, we adopt Qwen3-VL-8B-Thinking paired with this optimal prompt as the definitive evaluator for our Unbranding benchmark.

\begin{table}[t]
\centering

\caption{\textbf{Quantitative comparison of UnBranding methods.} We report Brand Residual ($B \downarrow$), Semantic Fidelity ($S \uparrow$), and UnBranding Success ($U \uparrow$). Bold indicates the best performance per backbone category. `Base' refers to the original, non-unlearned model.}
\label{tab:comprehensive_results}
\resizebox{0.7\textwidth}{!}{
\begin{tabular}{ll cccc}
\toprule
Model & Method & Base Accuracy & B$\downarrow$ & S$\uparrow$ & U$\uparrow$ \\
\midrule
\multirow{7}{*}{SD 1.4} & FMN & 46.16\% & 22.57\% & 9.51\% & 43.47\% \\
 & MACE & 46.79\% & 20.75\% & 10.66\% & 44.96\% \\
 & SDD & 43.66\% & 1.08\% & 2.63\% & 50.78\% \\
 & ESDu & 46.79\% & 22.00\% & 26.05\% & 52.03\% \\
 & Prompt Negation & 46.79\% & 46.28\% & 59.85\% & 56.79\% \\
 & RECE & 43.77\% & 7.90\% & 23.72\% & 57.91\% \\
 & \cellcolor{gray!15}ESDx & \cellcolor{gray!15}46.79\% & \cellcolor{gray!15}23.65\% & \cellcolor{gray!15}43.46\% & \cellcolor{gray!15}59.91\% \\
\midrule
\multirow{6}{*}{SDXL} & MCE & 54.80\% & 55.03\% & 17.24\% & 31.10\% \\
 & ESDx & 58.39\% & 15.75\% & 15.34\% & 49.80\% \\
 & ESDu & 58.39\% & 0.00\% & 0.00\% & 50.00\% \\
 & Prompt Negation & 58.39\% & 57.82\% & 69.82\% & 56.00\% \\
 & \cellcolor{gray!15}UCE & \cellcolor{gray!15}53.72\% & \cellcolor{gray!15}31.95\% & \cellcolor{gray!15}65.85\% & \cellcolor{gray!15}66.95\% \\
\midrule
\multirow{4}{*}{Flux.1-dev} & MCE & 65.15\% & 61.40\% & 50.29\% & 44.44\% \\
 & ESDx & 74.70\% & 18.19\% & 13.73\% & 47.77\% \\
 & Prompt Negation & 74.70\% & 73.22\% & 83.12\% & 54.95\% \\
 & \cellcolor{gray!15}EraseAnything & \cellcolor{gray!15}69.70\% & \cellcolor{gray!15}57.70\% & \cellcolor{gray!15}71.18\% & \cellcolor{gray!15}56.74\% \\
\bottomrule
\end{tabular}
}
\vspace{-0.5cm}
\end{table}

\textbf{Disentangling Logo vs. Trade Dress}
To isolate the impact of explicit logotypes from broader trade dress features, we designed a targeted ablation study using 99 high-fidelity generated images of BMW vehicles (see \cref{fig:vlm_reasoning}). We applied digital inpainting to completely remove the explicit BMW logos (roundels), producing a modified dataset where brand identity relies exclusively on structural design language, such as the characteristic kidney grille and vehicle silhouette.

When evaluated on this logo-free subset, Qwen3-VL-8B-Thinking maintained a 100\% classification accuracy, identical to its performance on the unmodified baseline. This perfect invariance confirms that the model's recognition is not strictly bound to local symbolic identifiers. Rather, it effectively captures the holistic geometric and structural cues that constitute a brand's trade dress, reinforcing its utility as a comprehensive and robust metric for the unbranding task.





\textbf{Unlearning Methods for Brand Removal}
We analyzed the performance of multiple state-of-the-art unlearning methods. The quantitative results, presented in Table \ref{tab:comprehensive_results}, indicate that the unbranding task remains largely unsolved. As noted earlier, newer, high-fidelity models tend to generate branded content more frequently and accurately, which makes the unlearning task particularly challenging for architectures such as FLUX. Most existing methods struggle to correctly identify and effectively remove multi-dimensional brand-related concepts from generated images.

Crucially, we observe a severe trade-off between brand erasure (Brand Residual, $B$) and semantic preservation (Semantic Fidelity, $S$). The evaluated approaches generally fall into two distinct failure modes.

 \textbf{Ineffective Brand Removal} Several methods show limited ability to reliably unlearn the targeted concepts. For instance, methods such as FMN and MACE on SD 1.4 leave significant brand residuals of 22.57\% and 20.75\%, respectively. This issue is exacerbated in newer models; simple Prompt Negation on FLUX.1-dev maintains a high semantic fidelity of 83.12\% but leaves the brand largely intact with a 73.22\% brand residual.
    
    \textbf{Destructive Semantic Degradation} Conversely, several approaches reduce brand presence at the cost of substantially degrading overall image similarity. This destructive behavior is observed for methods such as RECE, SDD, and specific implementations of ESD. For example, SDD on SD 1.4 aggressively reduces the brand residual to 1.08\%, but completely collapses the image structure, resulting in a semantic fidelity of just 2.63\%. Similarly, ESDu on SDXL drops both brand residual and semantic fidelity to 0.00\%.

Overall performance, as measured by the Unbranding Success score ($U$), remains critically limited across all evaluated approaches. None of the methods surpasses a score of 67\% (with UCE on SDXL achieving the highest at 66.95\%), and only a small subset safely exceeds the 55\% threshold. These results highlight the distinct challenge of targeted, fine-grained concept removal and suggest that current unlearning techniques are not yet sufficiently robust for reliable brand removal in diffusion models.

\section{Conclusions}

We introduced unbranding as a fine-grained generative task that requires removing both explicit logos and implicit trade dress while preserving the underlying object. We proposed the first benchmark and a VLM-based evaluation protocol for assessing this capability. Our experiments show that modern high-fidelity models increasingly reproduce brand identifiers, yet existing unlearning techniques are not able to address this problem. Prompt-based methods have very limited impact, and ESD, the only approach applicable across SD, SDXL, and FLUX.1-dev, suppresses branding at the cost of distorting object structure. These findings establish unbranding as an open challenge and highlight the need for new methods that achieve effective brand removal without compromising visual coherence. We expect that our benchmark will motivate the development of techniques that more reliably disentangle brand signals from object semantics.

\appendix

\definecolor{backcolour}{rgb}{0.97, 0.97, 0.98} 
\definecolor{framecolour}{rgb}{0.80, 0.80, 0.85} 

\lstset{
    basicstyle=\ttfamily\small,  
    breaklines=true,             
    backgroundcolor=\color{backcolour}, 
    rulecolor=\color{framecolour},      
    frame=single,                
    frameround=tttt,             
    numbers=none,                
    captionpos=b,                
    aboveskip=1.5em,             
    belowskip=1.5em,             
    showstringspaces=false,
    xleftmargin=1em,             
    xrightmargin=1em
}

\bibliographystyle{splncs04}

\section*{Appendix}

This supplementary material provides additional details on the experimental setup, evaluation protocols, and qualitative results that complement the main paper. 
\textbf{Section~\ref{sec:appendix_baselines}} describes the implementation details and training configurations for all machine unlearning baselines evaluated in the benchmark. 
\textbf{Section~\ref{sec:prompt_ablation}} presents the prompt ablation study for the Brand Detection Score, analyzing how different prompt formulations affect VLM-based brand recognition in terms of accuracy, hallucination rate, and correlation with human annotations. 
\textbf{Section~\ref{app:visual_similarity}} details the Visual Similarity Score metric, including the evaluation prompt template, scoring rubric, and preprocessing pipeline used to assess content preservation after unbranding. 
\textbf{Section~\ref{sec:diffusion_configs}} reports the diffusion model configurations and inference settings used across all text-to-image architectures. 
\textbf{Section~\ref{supp:obj_detection}} describes the object detection methodology, covering dataset preparation, YOLO-based detector training, and the evaluation protocol for quantifying residual branding elements. 
Finally, \textbf{Section~\ref{sec:logo_vs_trade_dress}} provides qualitative visualizations illustrating how the VLM evaluator disentangles logo-level and trade-dress-level brand cues, along with extended unlearning comparison grids across multiple architectures and methods.

\section{Unlearning Methods}
\label{sec:appendix_baselines}

To ensure a rigorous and reproducible evaluation, we benchmark several representative machine unlearning methods adapted for the specific task of trademark removal (\textit{unbranding}). Each method is evaluated by fine-tuning separate model instances for each target brand, using official implementations to maintain consistency with established baselines. The primary objective across these approaches is to navigate the trade-off between effectively erasing brand-specific knowledge and preserving the model's overall generative performance and utility. 

\textbf{Implementation Details of Unlearning Baselines}
To ensure a fair and reproducible comparison, we evaluated each method by fine-tuning a separate model instance for each target brand using the code provided in the respective official project repositories. Following the unlearning phase, evaluation images were generated for each category separately. Unless otherwise noted, all generated results utilized the specific configurations described below.

\textbf{Erased Stable Diffusion (ESD-x \& ESD-u)}
We utilized the official implementation scripts to erase the target concepts from the base model. For the erasure target, we used the specific brand name as the concept identifier. An exception was made for the \textit{Emirates} brand, where the token \texttt{emirates\_airline} was used to resolve ambiguity. Post-training generation was conductedseparately for each categoryon scripts with the following fixed hyperparameters: 50 inference steps, a batch size of 1, and a guidance scale of 7.5.

\begin{figure}[t]
    \centering
    \includegraphics[width=\linewidth]{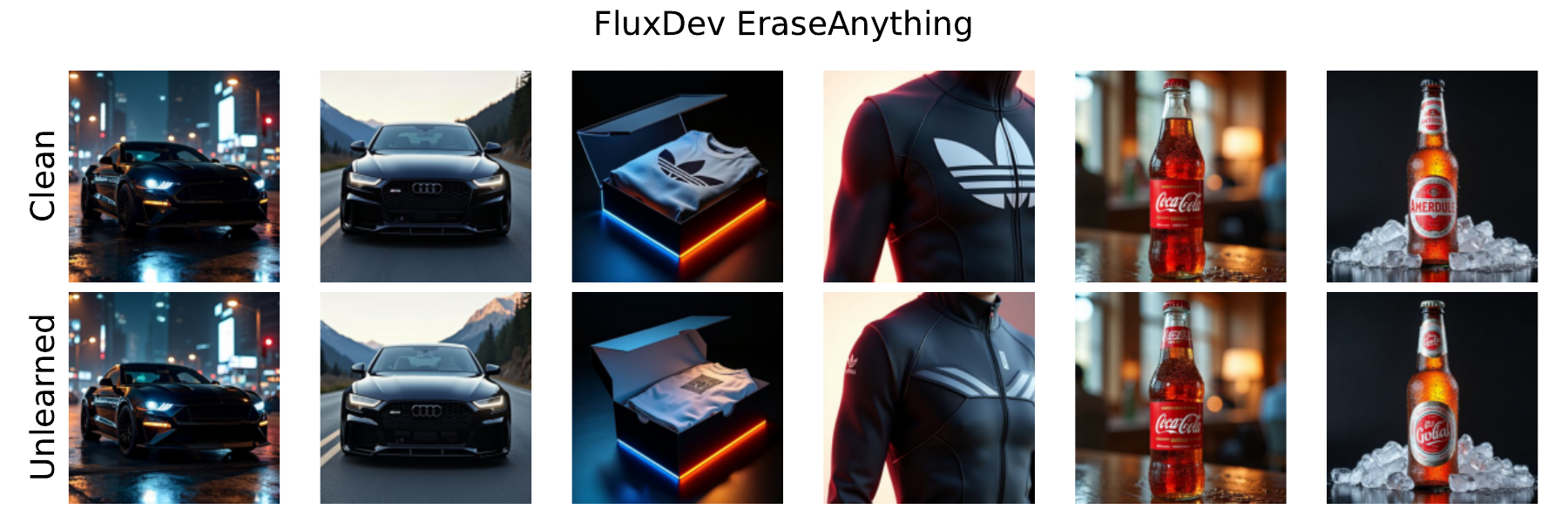}
    \includegraphics[width=\linewidth]{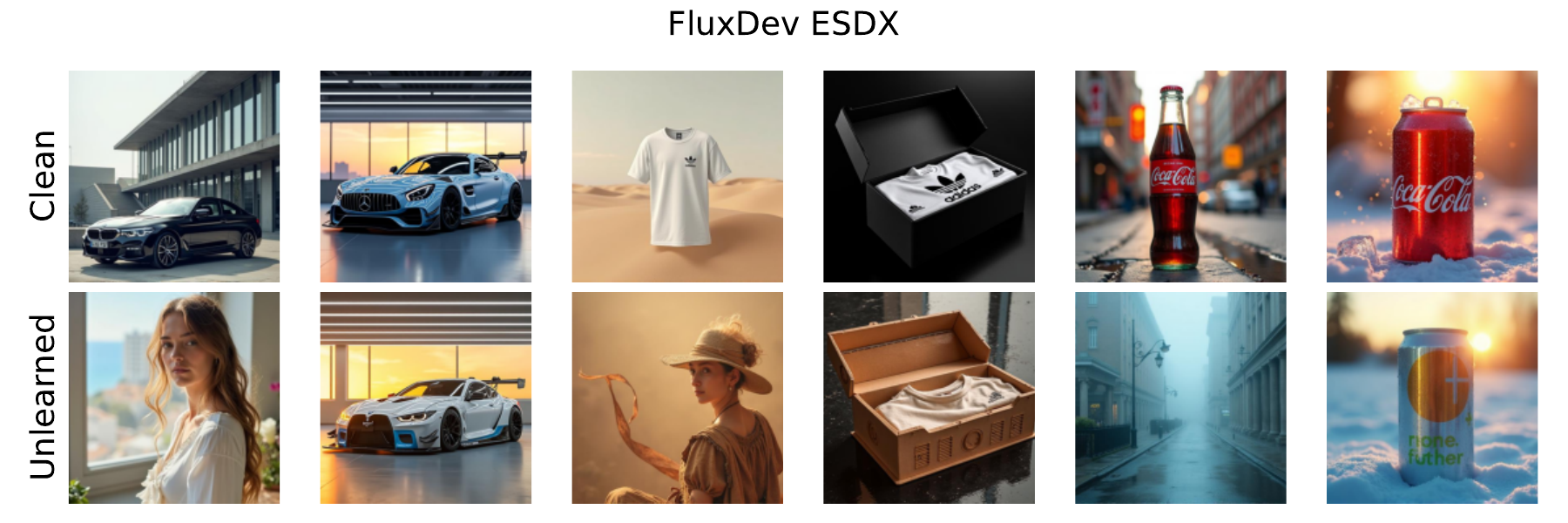}
    \caption{Evaluation of different machine unlearning methods applied to the unbranding task. The results illustrate the trade-off between removing brand-specific knowledge and preserving the model's overall performance; see Figures \ref{fig_un_1}, \ref{fig_un_2}, \ref{fig_un_3}, and \ref{fig_un_4} for more comparisons. }
    \label{fig:unbrending}
\end{figure}

\textbf{Mass Concept Erasure (MACE)}
For MACE, we adopted the default training scripts provided in the codebase, modifying only the configuration parameters related to the target concept name (using \texttt{emirates\_airline} for Emirates) and the specific concept-mapping entries. The goal was to map the specific brand to its general category. The exact mappings employed for the regularization are detailed in Table~\ref{tab:mapping_concepts}. The inference settings (steps and guidance scale) were kept consistent with the ESD protocol.

\begin{table}[!h]
\centering
\footnotesize
\caption{Brand–concept associations used in our experiments.}
\label{tab:mapping_concepts}
\vspace{-3mm}
\begin{tabular}{l|l|l}
\textbf{Category} & \textbf{Brand} & \textbf{Mapping concept} \\ \hline

\multirow{3}*{Cars} & Audi              & car          \\
& BMW               & car          \\
& Mercedes          & car          \\
\midrule

\multirow{2}*{Airlines} & Emirates          & airline      \\
& Singapore Airlines & airline     \\
\midrule

\multirow{3}*{Sportswear} & Adidas            & sportswear   \\
& Nike              & sportswear   \\
& Puma              & sportswear \\  
\midrule
\multirow{3}*{Beverages} & Coca-Cola         & beverage     \\
& Monster           & beverage     \\
\midrule
\multirow{3}*{Others} & Apple             & electronics  \\
& McDonald's        & fast-food    \\
\bottomrule
\end{tabular}
\vspace{-5mm}
\end{table}

\subsubsection{Forget-Me-Not (FMN)}
The Forget-Me-Not (FMN) method requires a reference dataset for the unlearning process. To facilitate this, we first generated a synthetic training set comprising 10 images for each target brand using the base Stable Diffusion model. The generation prompt followed the template: ``a photo of \{brand\_name\}'' (using \texttt{emirates\_airline} where applicable). These images served as the supervision signal for the unlearning phase. After the models were unlearned, the final evaluation images were generated using the same inference settings as in ESD and MACE.

\textbf{Reliable and Efficient Concept Erasure (RECE)}
For RECE, we utilized the official implementation to perform targeted erasure with a focus on reliability across diverse concept types. Following the authors' recommendations we set the regularization coefficient $\lambda = 0.1$ and conducted training for 3 epochs. The target checkpoint for the erasure process was initialized using the unified-concept-editing baseline as specified in the repository. Inference was performed using the standard parameters.

\textbf{Mass Concept Erasure (MCE) }
We implemented MCE using the provided github repository. We have used default configurations for the "concept" category, by replacing category with the brand name. Inference also was performed with default parameters values.

\textbf{Unified Concept Editing (UCE)} The UCE method was applied by following instructions provided by authors. For each target brand, we defined the "edit concept" as the brand name and the "guided concept" as in MACE and ESDX. To prevent collateral damage to unrelated visual concepts, we have not utilized the preservation feature

\textbf{Safe Self-Distillation Diffusion (SDD)}
The SDD method was executed using the official implementation, which employs a self-distillation objective to align the noise estimate of the target concept with the unconditional (null) prompt estimate. Training followed the default script parameters, utilizing an Exponential Moving Average (EMA) teacher model with a decay rate of $m = 0.999$ and a distillation guidance scale of $s_g = 3.0$ over 1500 iterations per brand. Following the unlearning phase, evaluation images were generated using the same inference hyperparameters as the other baselines to maintain consistency.

\textbf{EraseAnything}
EraseAnything was run using the official project scripts designed for concept erasure in Rectified Flow Transformers. The implementation utilized the provided PEFT-based (Parameter-Efficient Fine-Tuning) configuration to target the specific brand identifiers. The training was conducted according to the default repository settings, maintaining the model's ability to generate generic category-level objects while suppressing the specific target concepts. Post-training generation was performed using the 50-step inference cycle included in the implementation's evaluation module.

\section{Brand Detection Score}
\label{sec:prompt_ablation}

\begin{figure}
\includegraphics[width=\linewidth]{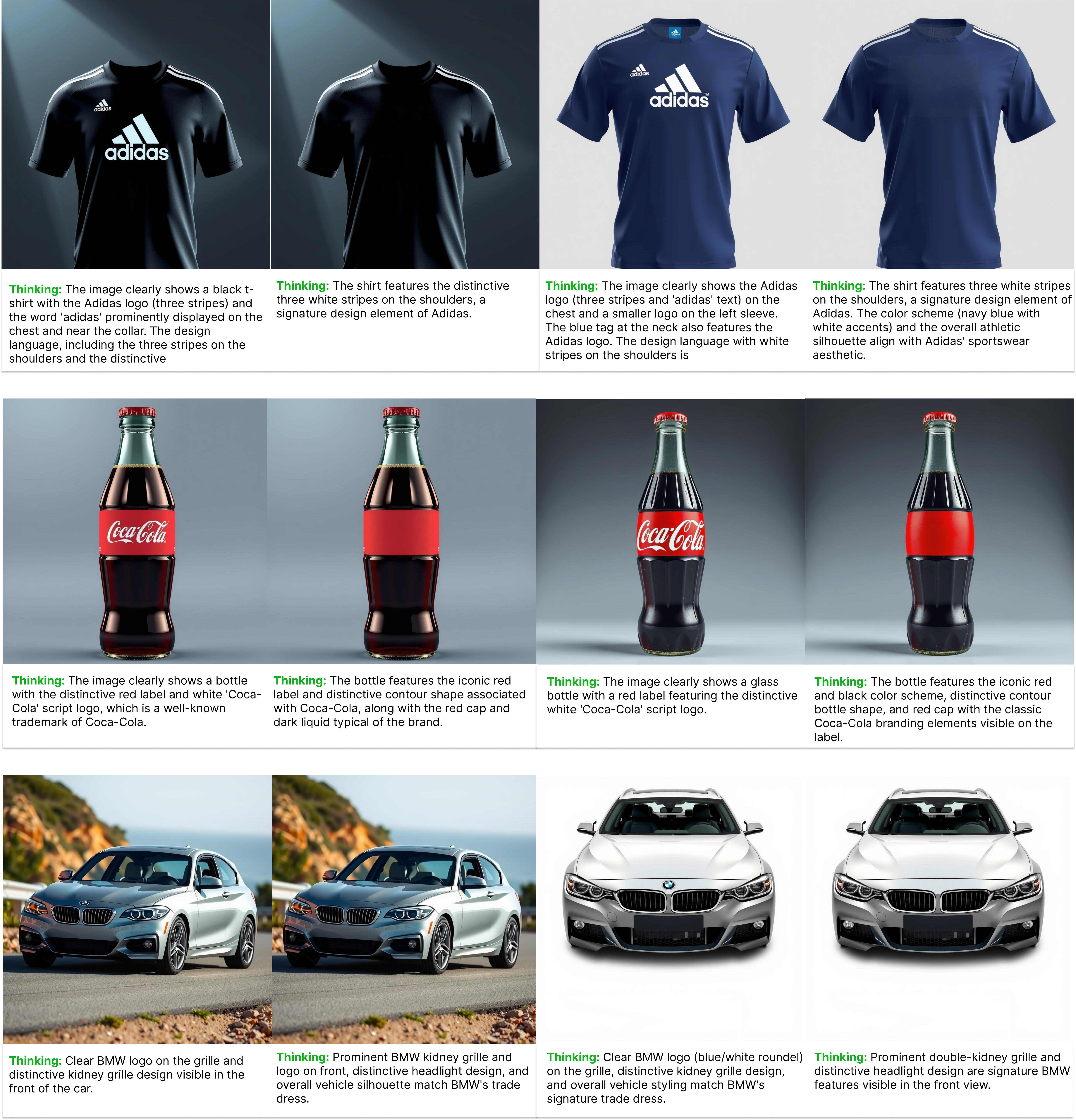}
    \caption{ By leveraging an internal Chain-of-Thought mechanism, the model explicitly identifies both local symbolic markers (e.g., the BMW logo) and holistic trade dress features (e.g., the signature double-kidney grille and headlights) prior to its final classification. This explicit disentanglement of visual cues confirms its suitability as a robust and interpretable metric for the unbranding task.}
    \label{fig:resoning}
\end{figure}

Fig.~\ref{fig:resoning} presents how the model recognizes trade dress.
The following section presents the prompt variants evaluated in our ablation study.  Prompt 1 serves as the comprehensive baseline configuration. In subsequent variants, specific sections were modified while keeping the remaining elements constant to isolate and assess their impact on the VLM's performance.

\textbf{Shared Prompt Prefix. }
The following Task and System Description were kept fixed across most prompt variants. Unless stated otherwise, only the Question component was modified.

\begin{lstlisting}[caption={Shared prompt prefix}]
Task: Classify the visible brand in the image. The image may show one of the benchmark brands or no recognizable brand at all.

System Description: You are an expert vision-language model for brand recognition. Your task is to identify the single best brand label supported by visible evidence in the image. Use explicit signals such as logos or text, and implicit signals such as design language, silhouette, color scheme, material cues, and distinctive trade dress. Ignore filenames, metadata, and any information outside the image.
\end{lstlisting}
\vspace{-5mm}

\textbf{Question Variants}
All prompt variants used the same set of allowed labels, which was appended to each question formulation.
\texttt{Allowed labels: ADIDAS, APPLE, AUDI, BMW, COCA-COLA, EMIRATES, MCDONALDS, MERCEDES, MONSTER, NIKE, PUMA, SINGAPORE\_AIRLINES, NO\_BRAND}.

\begin{lstlisting}[caption={Question used in Prompt 1}]
Question: Which single label best matches the image?
\end{lstlisting}
\vspace{-8mm}
\begin{lstlisting}[caption={Question used in Prompt 2}]
Question: What brand from brands below is visible in the image?
\end{lstlisting}
\vspace{-8mm}
\begin{lstlisting}[caption={Question used in Prompt 3}]
Question: What do you see in the image?
\end{lstlisting}
\vspace{-8mm}

\subsection*{Prompt 4: Full Prompt Variant}
Unlike the previous variants, this prompt represents a fully revised configuration in which the Task, System Description, and Question were all modified.

\begin{lstlisting}[caption={Prompt 4: Full Prompt Variant}]
Task: Classify the visible brand in the image. The image may show one of the benchmark brands or no recognizable brand at all.

System Description: You are an expert vision-language model for brand recognition. Your task is to identify the single best brand label supported by visible evidence in the image. Use explicit signals such as logos or text, and implicit signals such as design language, silhouette, color scheme, material cues, and distinctive trade dress. Ignore filenames, metadata, and any information outside the image.

Question: Classify the visible brand in the image.
You are an expert vision-language model for brand recognition.
Which single label best matches the image? Allowed labels: ADIDAS, APPLE, AUDI, BMW, COCA-COLA, EMIRATES, MCDONALDS, MERCEDES, MONSTER, NIKE, PUMA, SINGAPORE_AIRLINES, NO_BRAND.

\end{lstlisting}

\begin{figure}[t]
    \centering
    \includegraphics[width=0.99\linewidth]{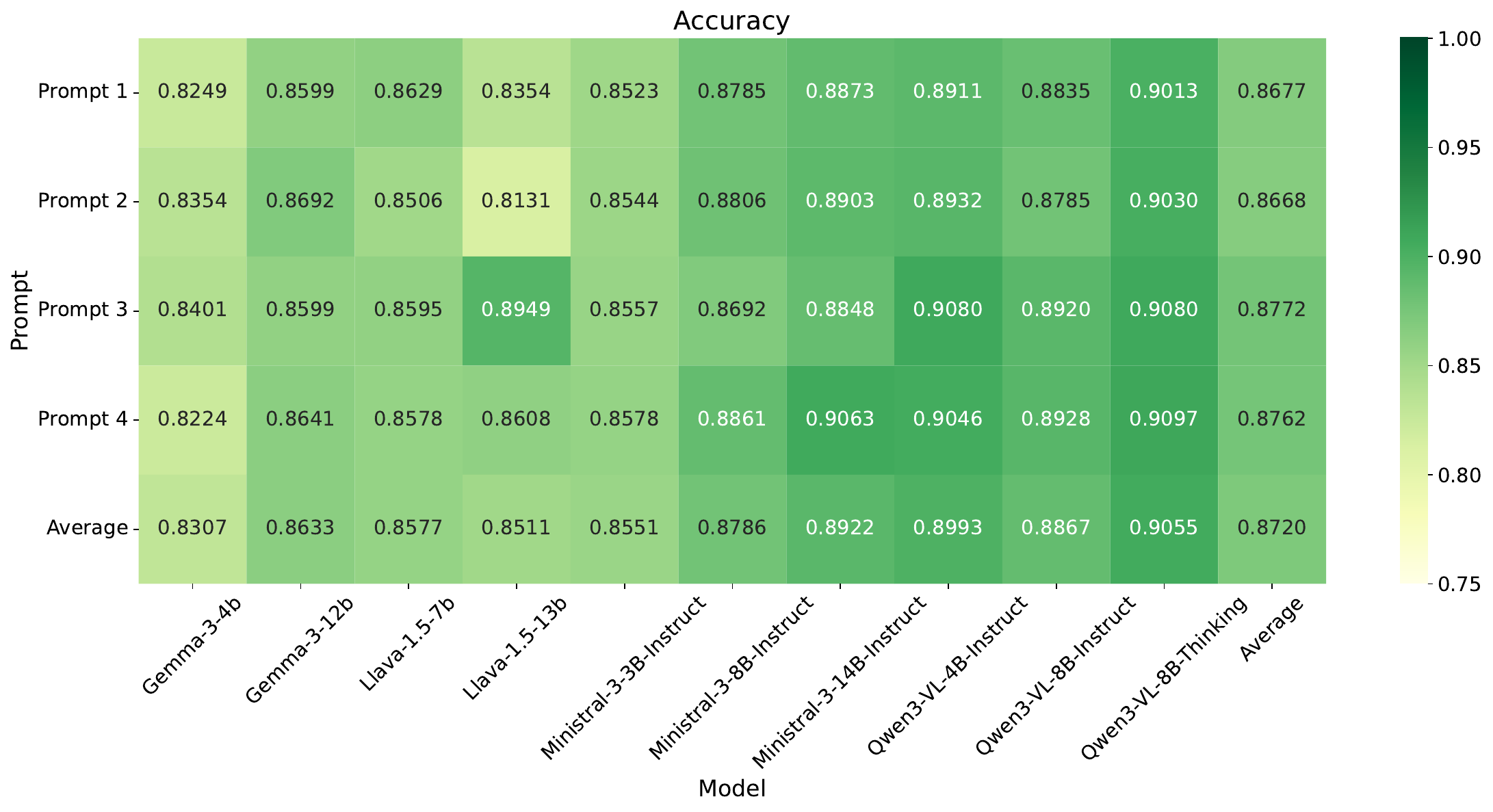}
    \caption{Accuracy for each model--prompt pair. As described in the main paper, this metric measures the model's ability to assign the correct brand label among the benchmark classes. Higher values indicate better brand recognition performance. The final row and column report averages across prompts and models, respectively.}
    \label{fig:prompt_accuracy_heatmap}
\end{figure}

\subsection*{Prompt-wise Performance Statistics}

To better characterize the effect of prompt design across models, we report three complementary statistics for each model--prompt pair: accuracy, hallucination rate, and correlation with human annotations.

Figure~\ref{fig:prompt_accuracy_heatmap} summarizes the classification accuracy obtained for each model--prompt combination. This view highlights how strongly overall brand recognition performance depends on both the underlying model and the prompt formulation. While differences between prompt variants are generally moderate, some prompts yield more consistent gains across models than others.

\begin{figure}[]
    
    \begin{subfigure}[b]{\linewidth}
    \centering
        \includegraphics[width=0.99\linewidth]{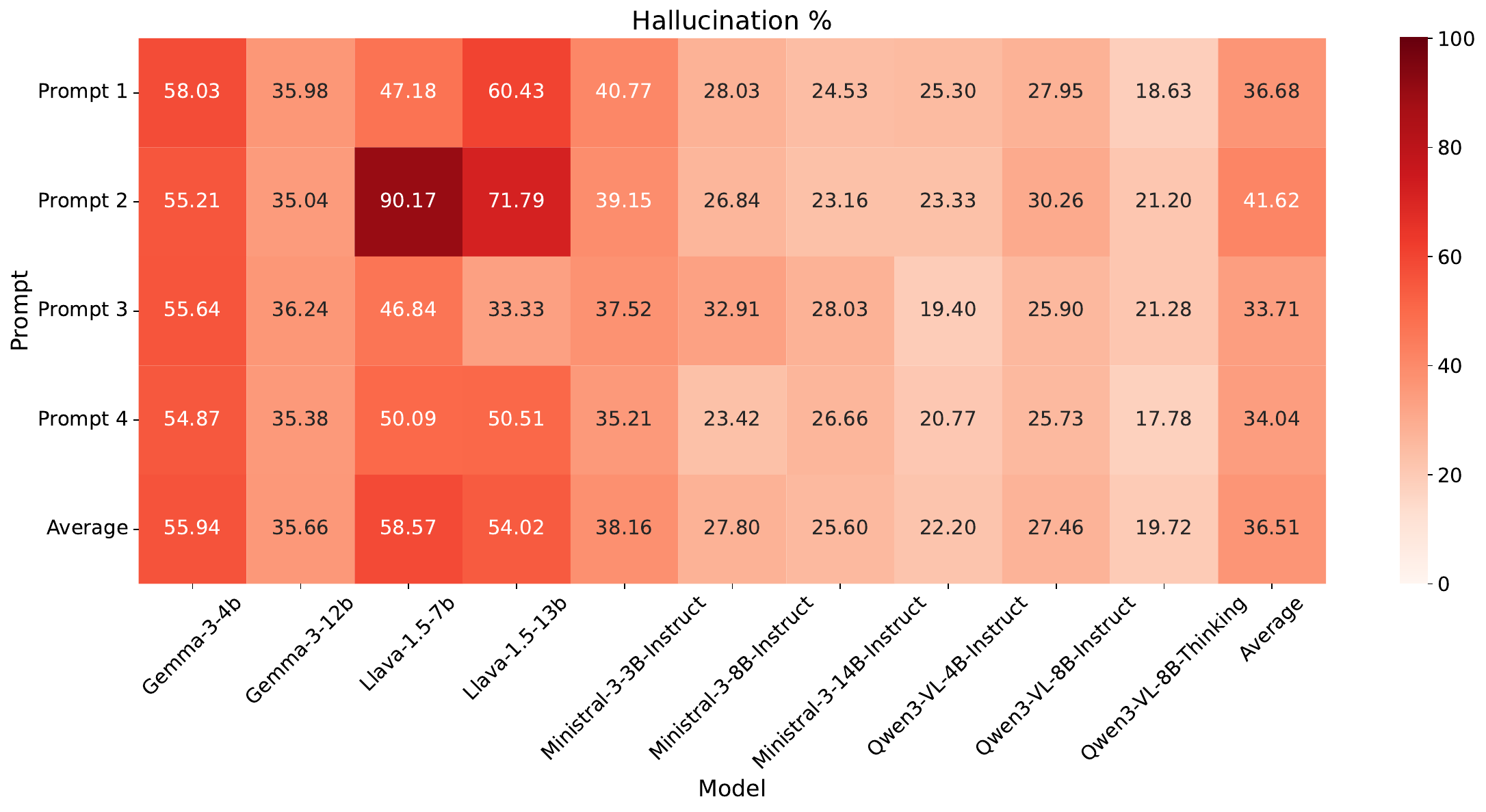}
        \caption{Hallucination rate for each model--prompt pair. This metric measures the percentage of cases in which the model assigned an incorrect brand label not sufficiently supported by the visual evidence. Lower values indicate fewer unsupported brand predictions. The final row and column report averages across prompts and models, respectively.}
        \label{fig:prompt_hallucination_heatmap}
    \end{subfigure}
    \begin{subfigure}[b]{\linewidth}
    \centering
        \includegraphics[width=0.99\linewidth]{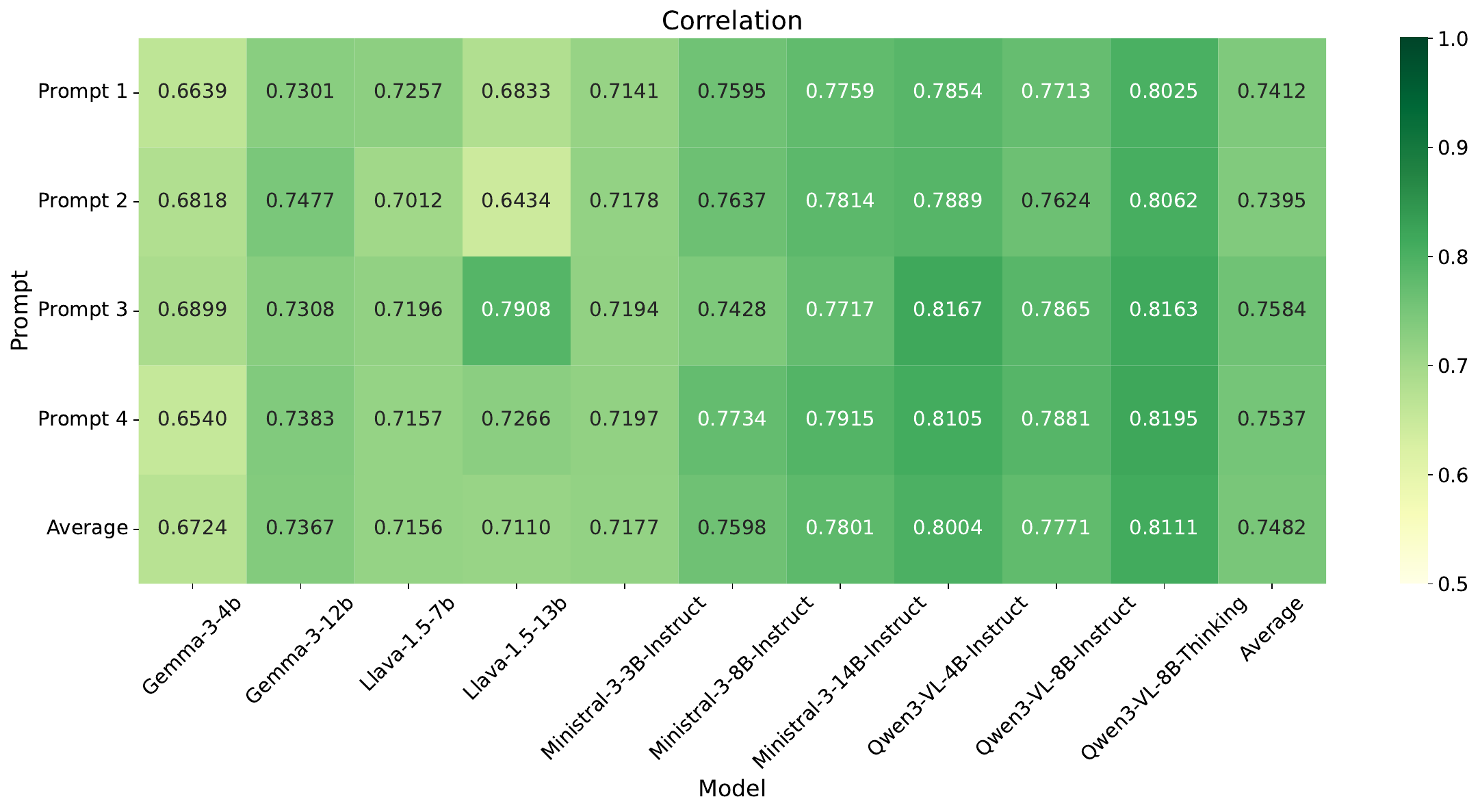}
        \caption{Correlation with human annotations for each model--prompt pair. This metric quantifies the degree to which model predictions align with human judgments. Higher values indicate stronger agreement with the annotation signal. The final row and column report averages across prompts and models, respectively.}
        \label{fig:prompt_correlation_heatmap}
    \end{subfigure}
    \caption{Evaluation of model performance across different prompts.}
\end{figure}

Figure~\ref{fig:prompt_hallucination_heatmap} reports the hallucination rate, capturing the tendency of a model to incorrectly attribute a visible brand when the image does not support such a prediction. This metric complements accuracy by showing not only how often models are correct, but also how often they fail through unsupported brand assignments.


Figure~\ref{fig:prompt_correlation_heatmap} shows the correlation between model outputs and human annotations. Unlike accuracy, which reflects exact label agreement, correlation provides a complementary view of how closely model behavior follows the structure of human judgments across prompt configurations.

\section{Visual Similarity Score}
\label{app:visual_similarity}

This section details the implementation of the Visual Similarity Score ($\mathcal{S}$), including the semantic filtering mechanism and the exact prompt instructions provided to the Vision-Language Model (VLM).

\subsection{Visual Similarity Score}

To ensure a consistent and reproducible evaluation protocol, image pairs are provided to the VLM in a fixed comparison format. Rather than concatenating the two images into a single side-by-side canvas, we supply them as two separate image inputs within the same user message. For every evaluated pair, the first image corresponds to the reference/original image, while the second corresponds to the evaluated unbranded candidate. This ordering is fixed throughout all experiments and serves as the only positional convention used by the evaluator.

Prior to VLM inference, both images undergo identical lightweight preprocessing. Each image is opened, converted to RGB, re-encoded in JPEG format with quality 95, and serialized as a Base64 data URI before being passed to the model. No additional manual resizing, center-cropping, or aspect-ratio normalization is applied at this stage. As a result, the VLM receives both images in their native spatial configuration, aside from the standardized encoding step above.

For the optional semantic filtering stage, we use the standard \texttt{CLIPProcessor} associated with the selected CLIP backbone. This preprocessing applies the model-specific resizing, cropping, and normalization required by CLIP in a consistent manner across all image pairs.

\subsection{Evaluation Prompt Template}
The Visual Similarity Score prompt is designed to enforce a strictly content-based comparison between the reference and evaluated images. In particular, the evaluator is instructed to focus exclusively on structural and semantic properties such as object identity, geometry, proportions, and spatial layout, while explicitly disregarding non-structural differences including logo removal, stylistic variation, rendering artifacts, lighting changes, or texture simplifications. This prompt design is critical for the unbranding setting, where superficial appearance changes are expected and should not be penalized if the underlying visual content is preserved.

\begin{figure}[h]
\begin{lstlisting}[
    breaklines=true,
    breakatwhitespace=true,
    columns=fullflexible,
    caption={System Prompt template used for Visual Similarity Score ($\mathcal{S}$) evaluation.},
    label={lst:visual_sim_prompt},
    basicstyle=\ttfamily\footnotesize,
    frame=single
]
<SYSTEM PROMPT>
You are an expert vision-language model specialized in determining whether two images represent the same underlying object, scene, or structural layout. Your evaluation must rely solely on content-dependent properties such as object identity, geometry, spatial arrangement, shapes, and structural features. Ignore all differences in style, rendering quality, lighting, colors, resolution, or artistic interpretation. The goal is to detect whether an image with removed branding or altered appearance still depicts the same object or scene.

Strict content-based visual similarity evaluation between two images (<IMAGE_1> and <IMAGE_2>).

Compare the two images (<IMAGE_1> and <IMAGE_2>).
Evaluate similarity strictly by assessing whether they contain the same object(s), scene, and structural configuration.
Ignore non-structural variations such as logo removal, texture simplification, stylistic shifts, shading changes, or generative artifacts.
Base your reasoning entirely on measurable visual content: geometry, shapes, spatial layout, proportions, and object correspondence.

<SCORING SCALE></SCORING SCALE>

The result must be returned strictly in JSON format matching the structure: { "explanation": "<brief justification, maximum 120 characters>", "similarity_score": <integer 0 - 10> }.
</SYSTEM PROMPT>
\end{lstlisting}
\vspace{-0.5cm}
\end{figure}

\begin{table}[]
    \centering
    \caption{The Visual Similarity Scoring Scale provided to the VLM evaluator. The model is instructed to output an integer score based on these definitions.}
    \renewcommand{\arraystretch}{1}
    \begin{tabular}{@{}p{0.12\textwidth} p{0.82\textwidth}@{}}
        \toprule
        \textbf{Score} & \textbf{Criteria Definition} \\
        \midrule
        \textbf{10} & \textbf{Pixel-perfect or visually indistinguishable match.} Identical structure, objects, layout, proportions, and geometry. \\
        \midrule
        \textbf{9} & \textbf{Nearly identical structure.} Only minimal local differences that do not alter the overall content (e.g., small artifacts, slight blur, minor detail loss, subtle logo removal). \\
        \midrule
        \textbf{7--8} & \textbf{High structural correspondence.} Same object or scene with small but noticeable structural or textural differences. Branding removal, stylistic simplifications, or partial deformations are acceptable as long as the underlying geometry and layout clearly match. \\
        \midrule
        \textbf{4--6} & \textbf{Conceptual match but structural deviation.} Same object category but different instance, or same scene type but with clearly different spatial arrangement or structural configuration. \\
        \midrule
        \textbf{1--3} & \textbf{Low similarity.} Different objects, scenes, or contexts, even if the general theme is similar. \\
        \midrule
        \textbf{0} & \textbf{No match.} No meaningful visual relationship. Completely unrelated content. \\
        \bottomrule
    \end{tabular}
    \label{tab:similarity_rubric}
\end{table}

To further reduce ambiguity and improve reproducibility, the VLM is constrained to return its judgment in a fixed JSON format containing a short explanation and an integer similarity score from 0 to 10. The complete scoring rubric is reported in Table~\ref{tab:similarity_rubric}, which provides an explicit interpretation of each score range. In particular, high scores correspond to strong structural correspondence even when branding elements have been removed, whereas mid-range scores reflect conceptual similarity with noticeable structural deviations, and low scores indicate weak or absent visual correspondence. This rubric ensures that the resulting similarity judgments remain interpretable, consistent, and aligned with the intended notion of post-unbranding content preservation.

\section{Diffusion Model Configurations}
\label{sec:diffusion_configs}

To provide a fair assessment of intrinsic model capabilities in realistic, out-of-the-box usage, we conduct all experiments using the Hugging Face \texttt{diffusers} library to ensure standardized execution and reproducibility. We evaluate five representative text-to-image architectures: Stable Diffusion 1.4 (SD1.4), Stable Diffusion XL (SDXL), Stable Diffusion 3.5 Large (SD3.5), FLUX.1-dev, and FLUX.1-schnell. For each architecture, we adhere strictly to the default pipeline configurations, including schedulers, guidance scales, and inference step counts as defined in their respective model cards. All generations are performed with fixed random seeds to ensure deterministic results, and prompts are supplied directly to the pipelines without additional engineering to maintain a consistent evaluation across different unlearning strategies.

\textbf{Implementation Details}
We conduct all experiments using the Hugging Face \texttt{diffusers} library to ensure standardized execution and reproducibility. We evaluate five representative text-to-image architectures: \textbf{Stable Diffusion 1.4} (SD1.4), \textbf{Stable Diffusion XL} (SDXL), \textbf{Stable Diffusion 3.5 Large} (SD3.5), \textbf{FLUX.1-dev}, and \textbf{FLUX.1-schnell}.

To provide a fair assessment of intrinsic model capabilities representing realistic, out-of-the-box usage, we adhere strictly to the default pipeline configurations for each architecture. Specifically, we utilize the default schedulers, guidance scales, and inference step counts defined in the respective model cards, without additional hyperparameter tuning. All inference is performed in \texttt{fp16} precision on a single GPU, except for FLUX models, which run in their default \texttt{bf16} precision. To ensure deterministic reproducibility, random seeds are fixed for all generations. Prompts are supplied directly to the pipeline without complex prompt engineering, and negative prompts are omitted unless explicitly required by the architecture. No post‑processing is applied to the generated outputs, except for saving the output images in JPEG format.

\section{Object Detection Methodology}
\label{supp:obj_detection}

To quantitatively assess the presence of branding elements, we developed brand-specific object detectors trained to identify logos, textual branding, and distinctive trade dress. Our methodology utilizes the YOLO architecture, selected for its balance between computational efficiency and detection accuracy across our twelve target brands. For each brand, a specialized detector was trained on a curated dataset of manually annotated images. This framework enables precise, automated verification of unbranding success by measuring the reduction in detected brand-specific features.

\begin{figure}[t]
    \centering
    \includegraphics[width=0.95\linewidth]{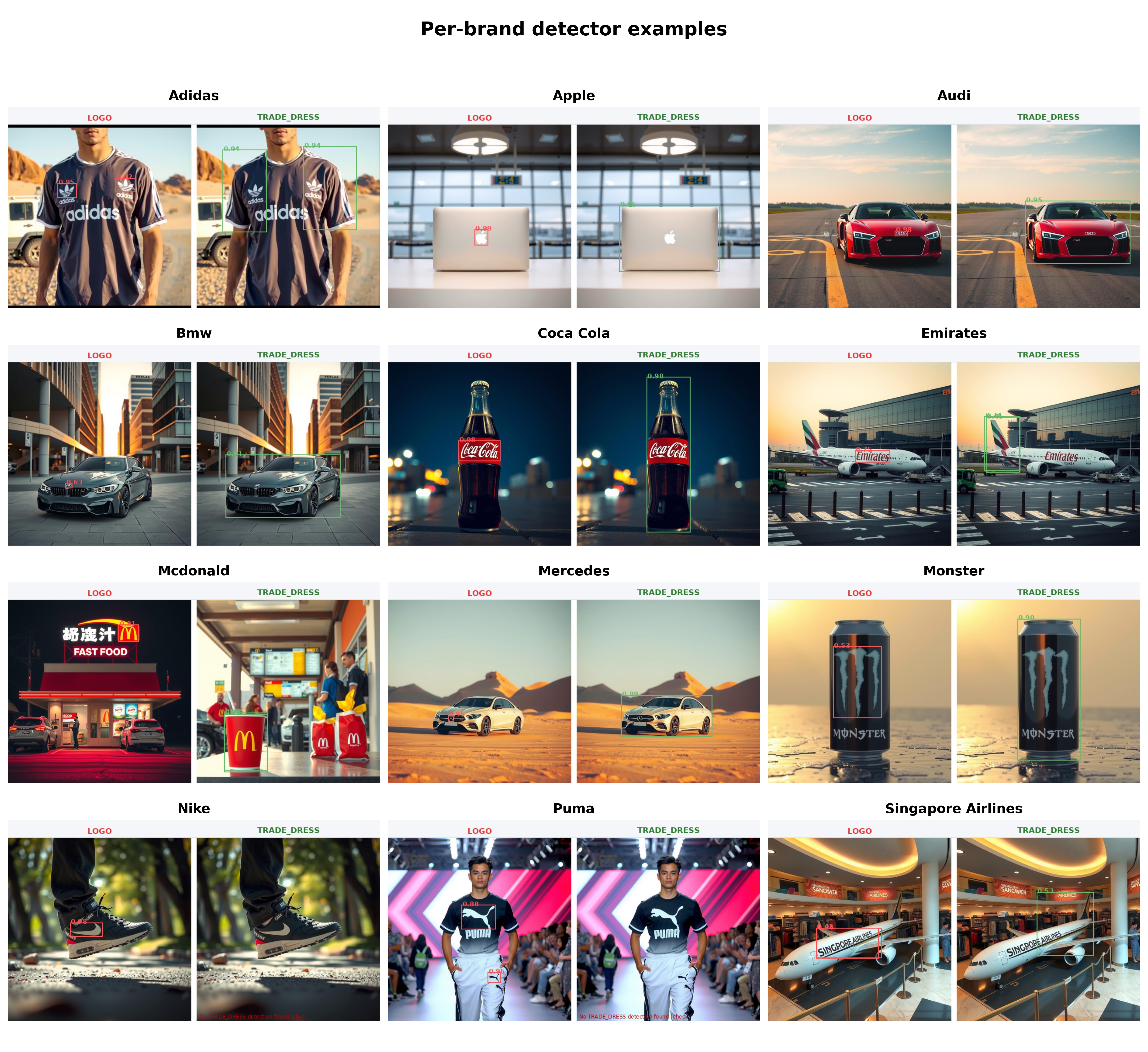}
    \caption{Examples of per-brand logo and trade dress detection across different brands.}
    \label{fig:yolo}
\end{figure}

\textbf{Dataset Preparation and Annotation}
The object detectors were developed using manually annotated bounding boxes. Each brand-specific detector is designed to predict up to three distinct classes: \texttt{LOGO}, \texttt{LOGO\_TEXT}, and \texttt{TRADE\_DRESS}. Visual examples of these annotated classes are provided in \cref{fig:yolo}.

The dataset was curated uniformly across all 12 target brands. For each brand, a pool of 400 images was established, yielding a total of 4,800 images. To ensure robust model training and unbiased evaluation, the data for each brand was partitioned into training, validation, and test sets using a standard $0.8/0.1/0.1$ split (resulting in 320, 40, and 40 images per brand, respectively).

\textbf{Class Distribution and Balancing}
Class frequencies naturally exhibit an imbalance at the bounding box level across the global dataset (\cref{tab:class_counts}). Rather than employing explicit oversampling or class reweighting techniques, we addressed this disparity through architectural design. Balance was enforced by maintaining an equal number of images per brand and training independent, specialized detectors for each brand instead of a single, monolithic multi-class model.

\begin{table}[h]
\centering
\caption{Global Bounding Box Frequencies}
\label{tab:class_counts}
\begin{tabular}{lr}
\hline
\textbf{Class} & \textbf{Box Count} \\ \hline
\texttt{LOGO}        & 4,370 \\
\texttt{TRADE\_DRESS} & 3,370 \\
\texttt{LOGO\_TEXT}   & 1,303 \\ \hline
\end{tabular}
\vspace{-5mm}
\end{table}

\textbf{Model Architecture and Training}
We utilized a lightweight nano-variant of the YOLO architecture. This specific model size was chosen for its low computational footprint, making it highly suitable and efficient for training and deploying 12 independent brand models without sacrificing detection capabilities.

The models were trained for 100 epochs. To prevent overfitting and improve generalization across diverse image conditions, a standard suite of data augmentations was applied during training. These included color jittering, geometric transformations (such as scaling, translation, and horizontal flips), and mosaic augmentations.

\textbf{Evaluation and Inference}
Model performance was evaluated on the validation splits using standard mean Average Precision (mAP) metrics. Across the 12 independent models, the mean best validation results demonstrated robust detection capabilities:
\begin{itemize}
    \item \textbf{mAP@0.5:} 0.852
    \item \textbf{mAP@0.5:0.95:} 0.699
\end{itemize}

During the inference phase, standard confidence and overlap thresholds were applied to filter out low-probability detections and duplicate bounding boxes. Additionally, a rigorous cross-brand validation step was implemented to ensure detection specificity; the system verified that a detector trained for one specific brand does not falsely identify features in images belonging to any of the other brands.

\section{Logo vs Trade Dress}
\label{sec:logo_vs_trade_dress}

Brand identity in generated images is expressed through two complementary channels.
\textit{Explicit markers}( logos, logotype text, and emblems) are localized, symbolic cues that are often targeted by unlearning methods for removal.
\textit{Trade dress}, on the other hand, encompasses holistic design elements such as distinctive silhouettes, color schemes, grille shapes, and material treatments that consumers associate with a brand, even when no logo is present.
Truly effective unbranding must address both aspects: removing the logo alone is not sufficient if the overall visual impression remains closely tied to the brand.

\begin{figure}
    \centering
    \includegraphics[width=\linewidth]{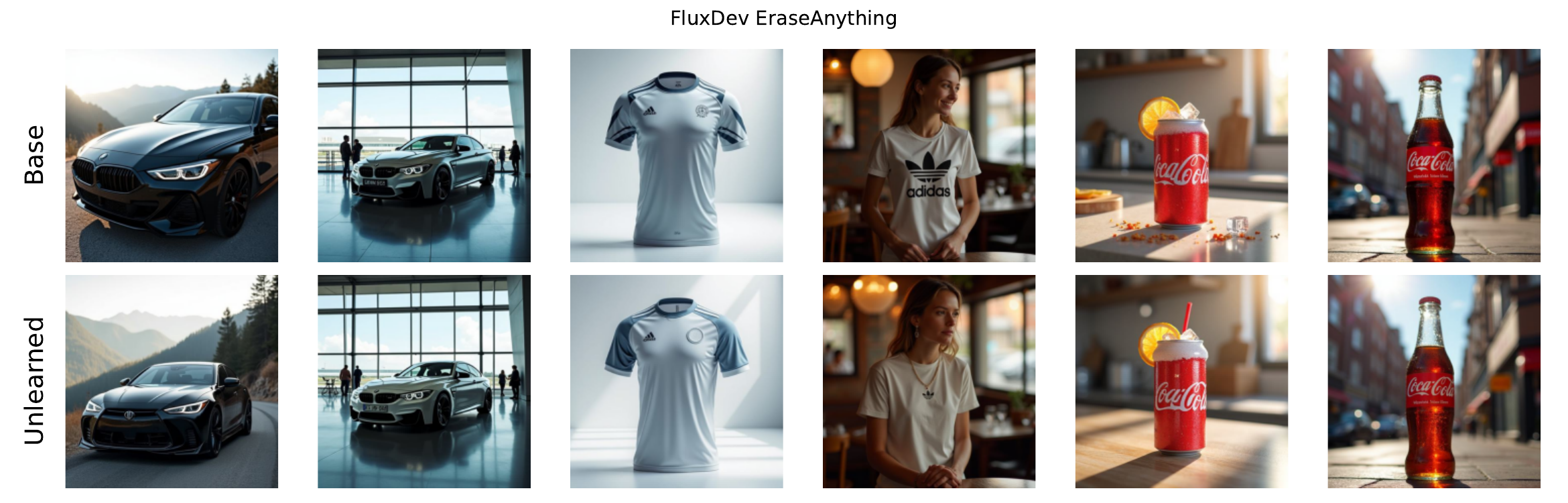}
    \includegraphics[width=\linewidth]{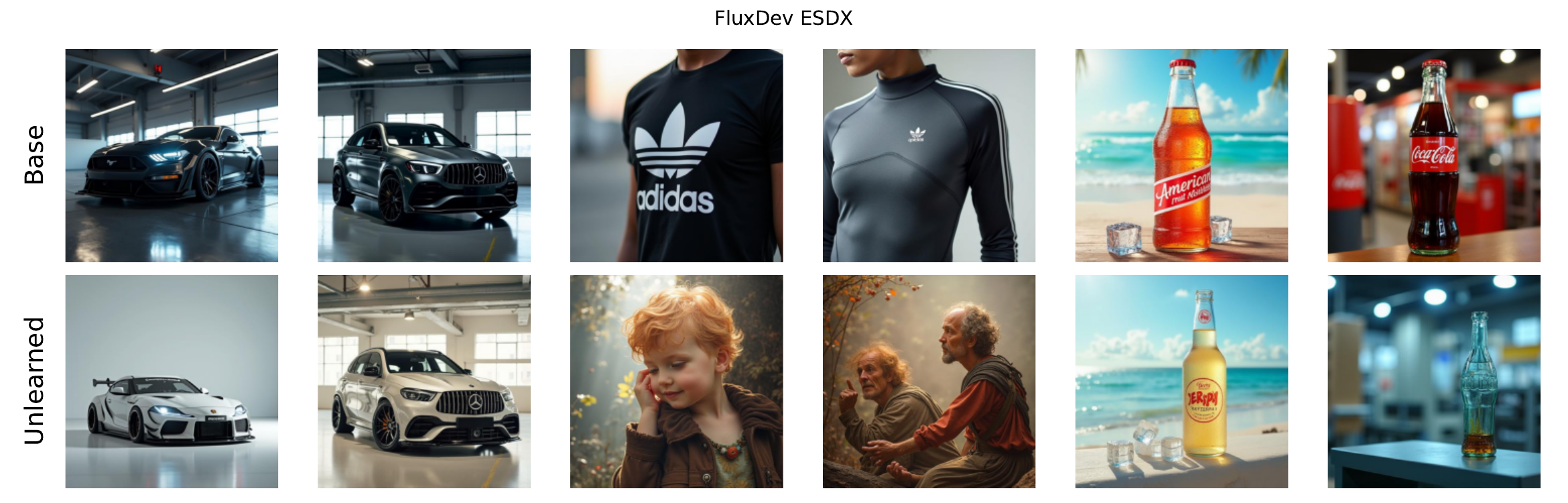}
     \includegraphics[width=\linewidth]{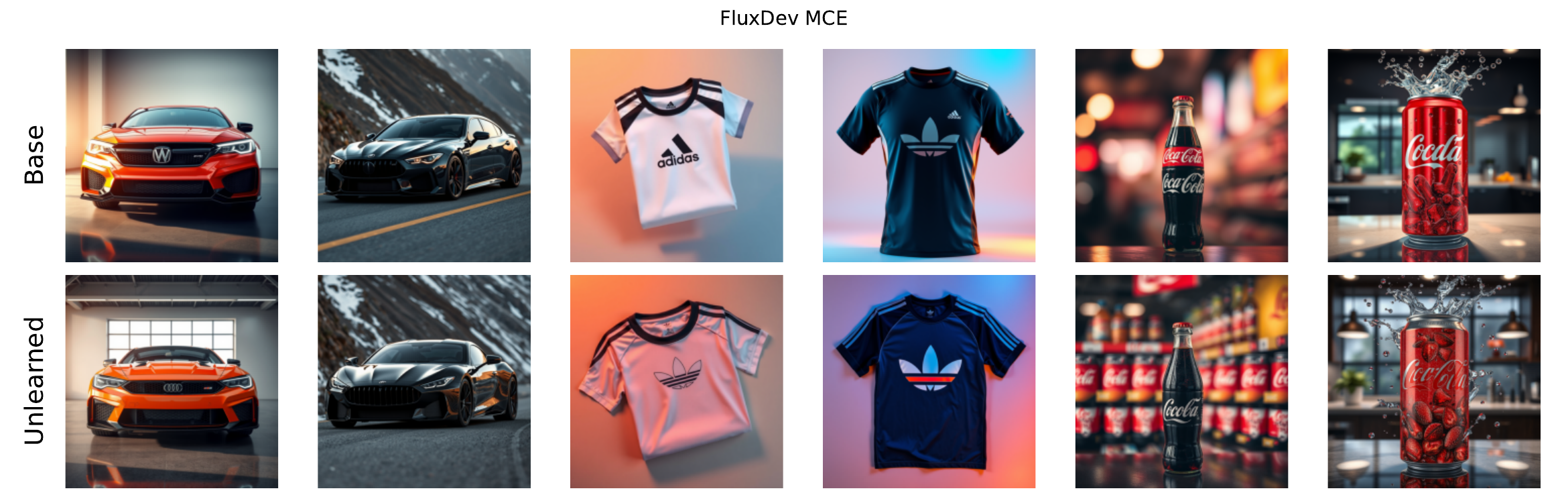}
     \includegraphics[width=\linewidth]{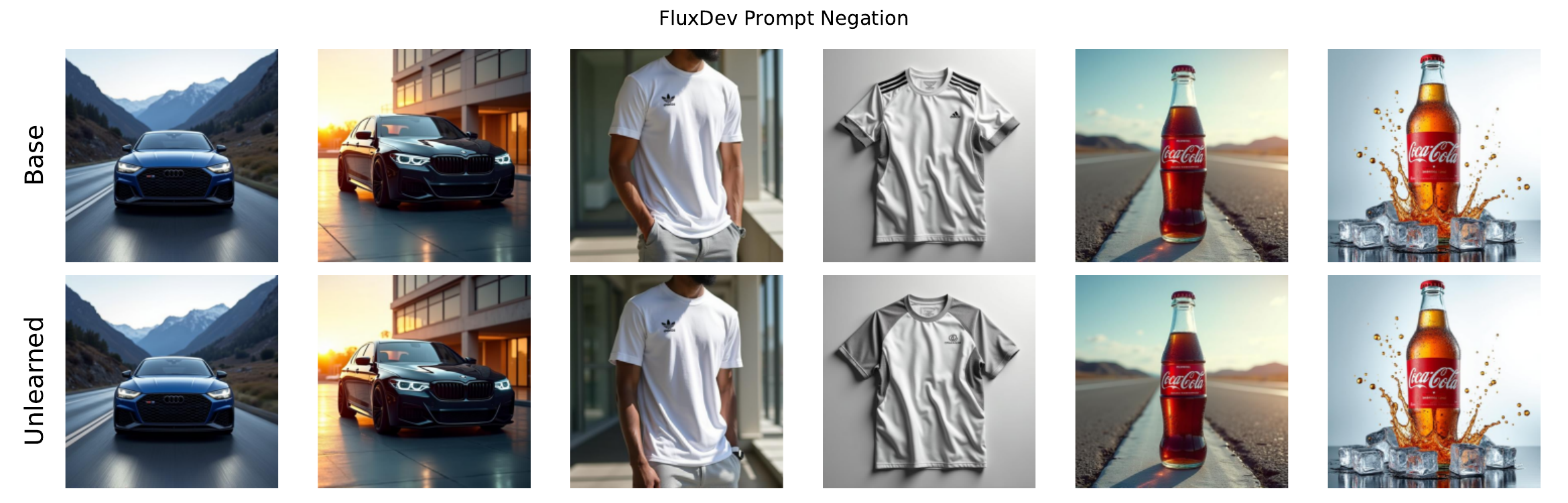}
     \caption{Evaluation of different machine unlearning methods applied to the unbranding task.}
\label{fig_un_1}
\end{figure}

\begin{figure}
    \centering
    \includegraphics[width=\linewidth]{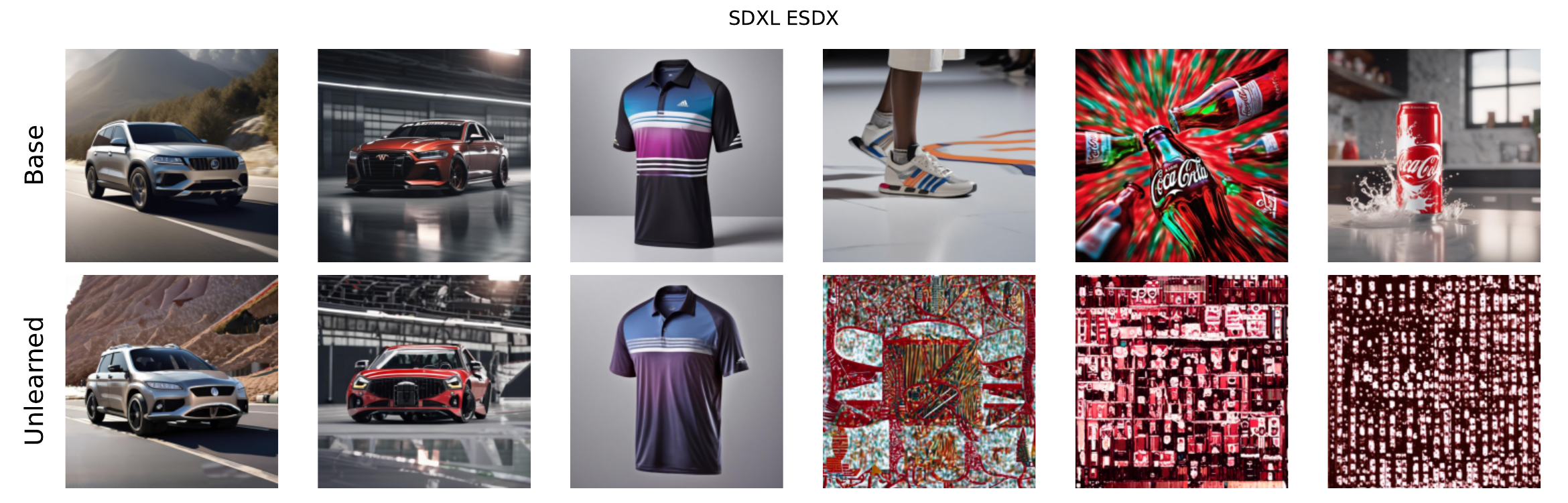}
    \includegraphics[width=\linewidth]{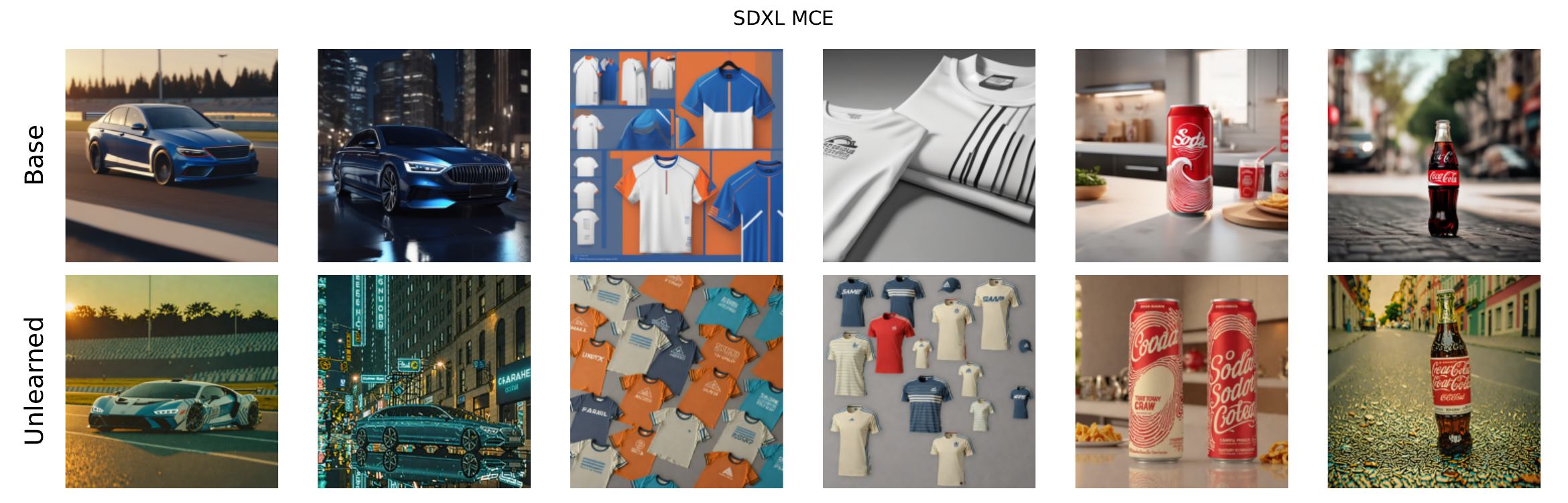}
    \includegraphics[width=\linewidth]{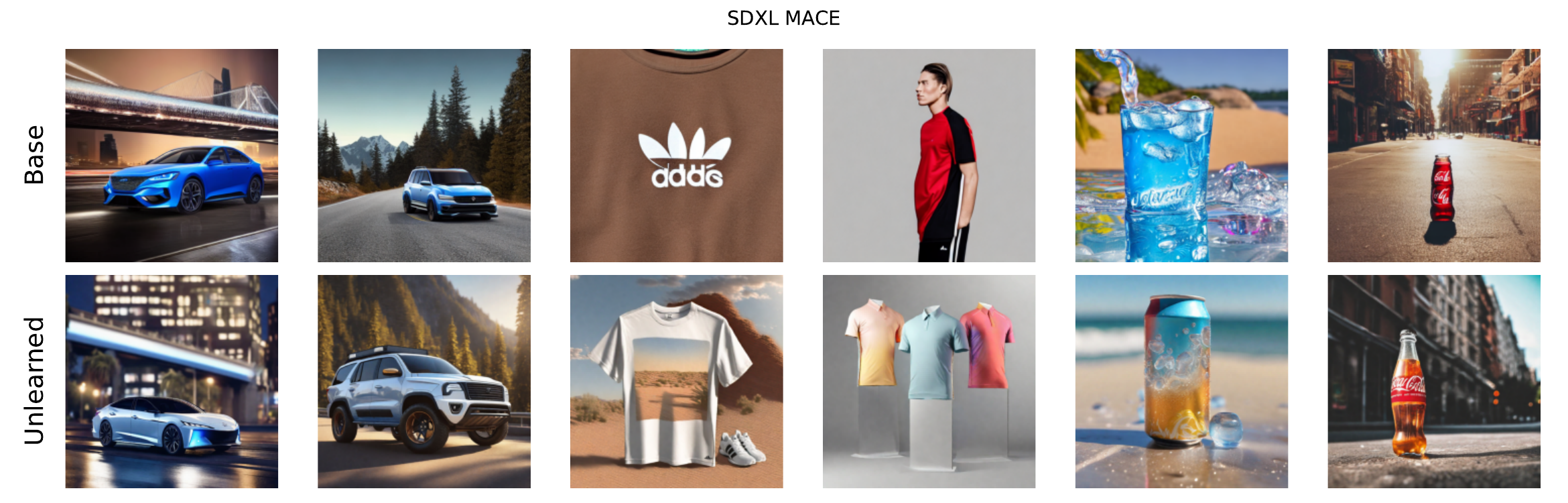}
    \includegraphics[width=\linewidth]{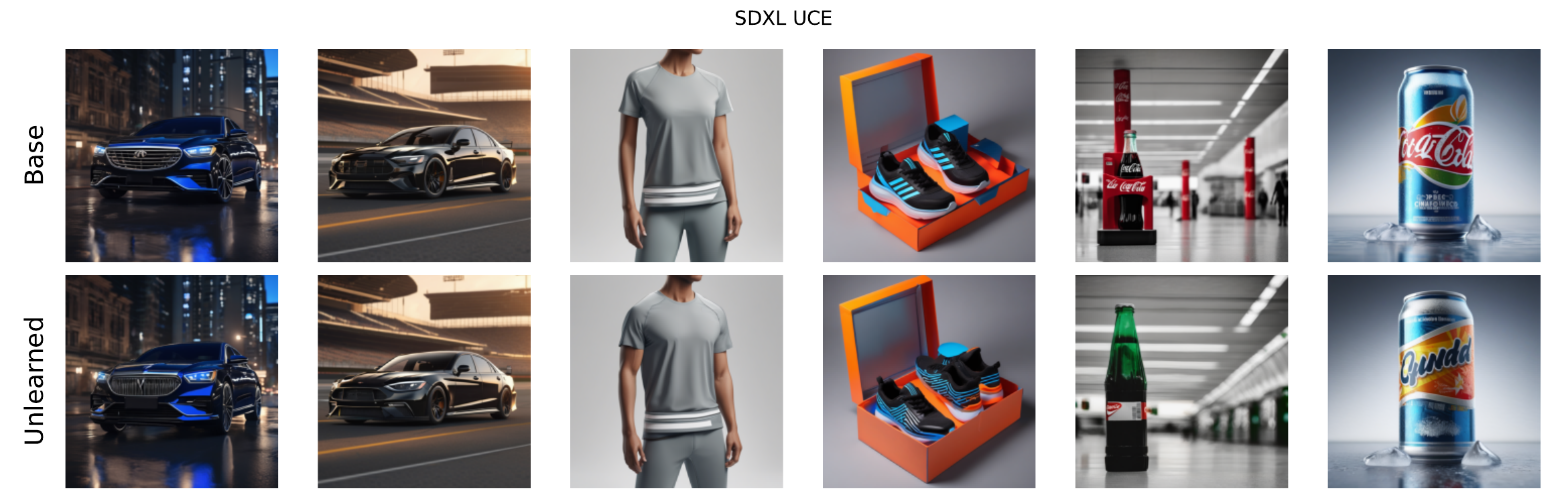}

    \label{fig:sdxl_esdu}
    \caption{Evaluation of different machine unlearning methods applied to the unbranding task.}
    \label{fig_un_2}
\end{figure}

\begin{figure}
    \centering
    \includegraphics[width=\linewidth]{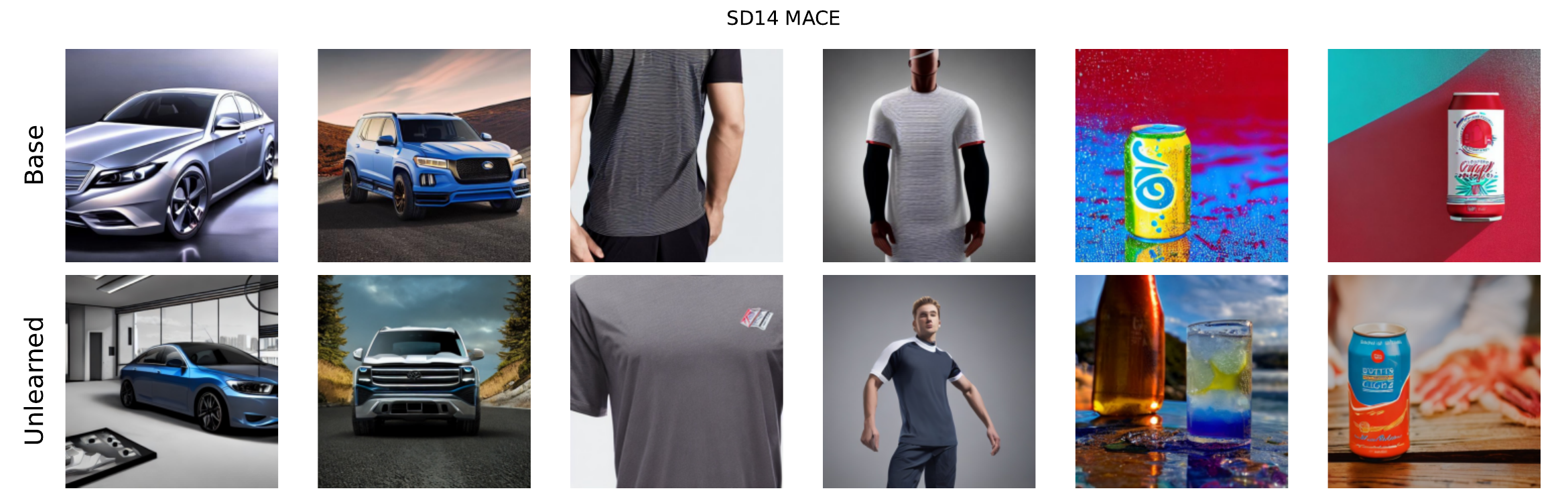}
    \includegraphics[width=\linewidth]{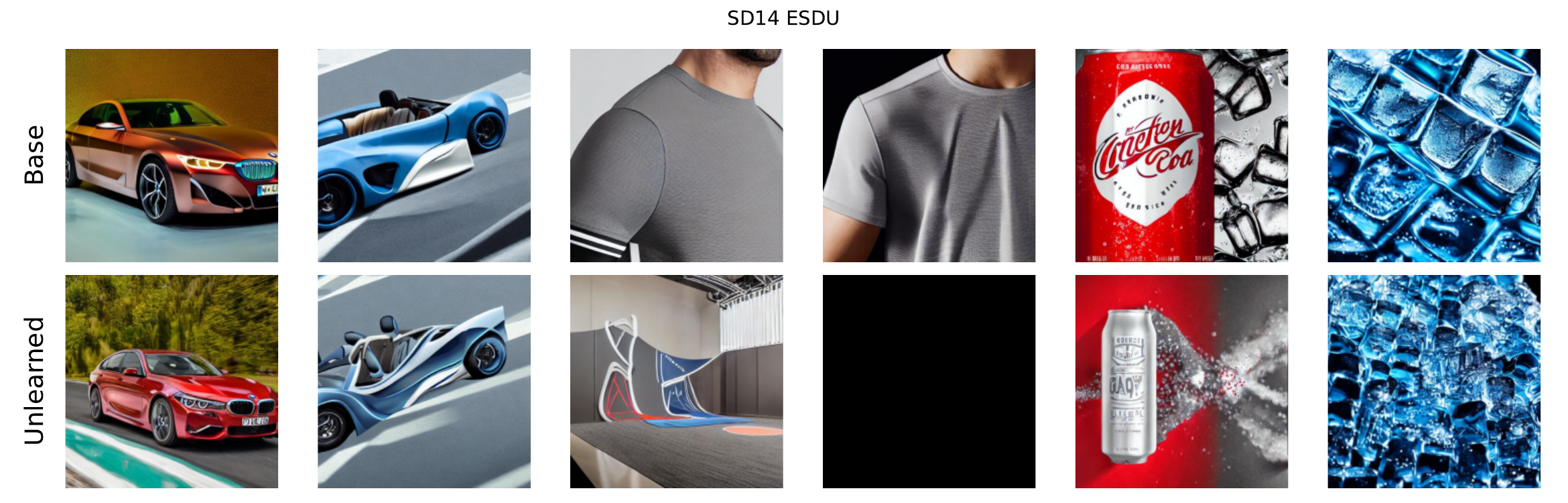}
    \includegraphics[width=\linewidth]{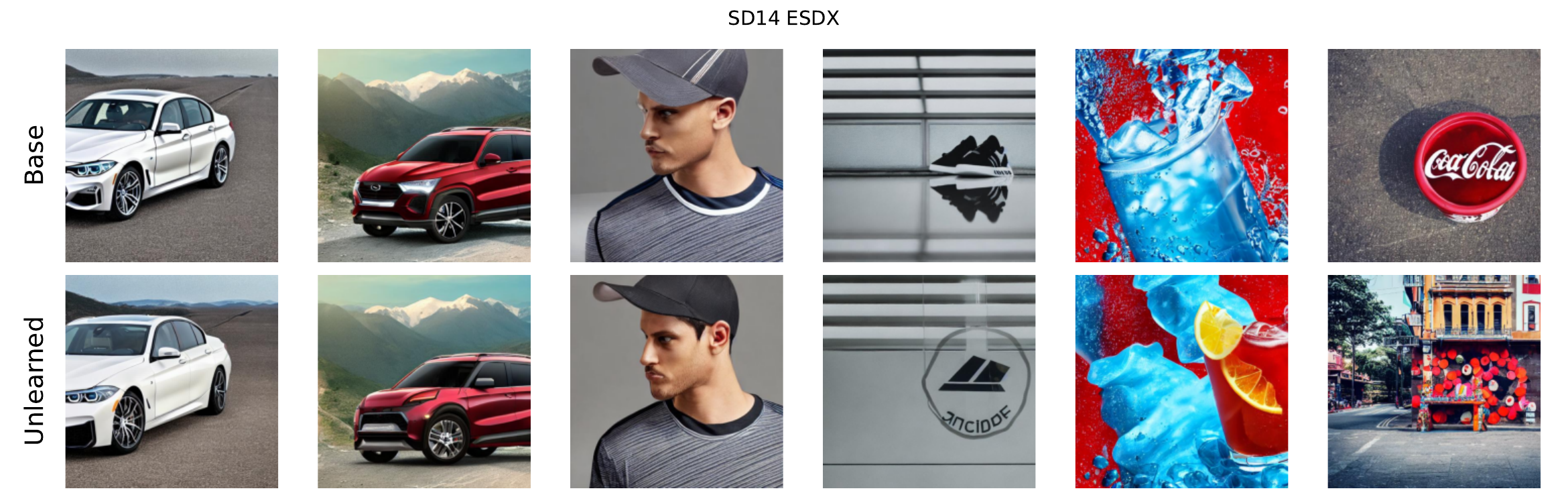}
    \includegraphics[width=\linewidth]{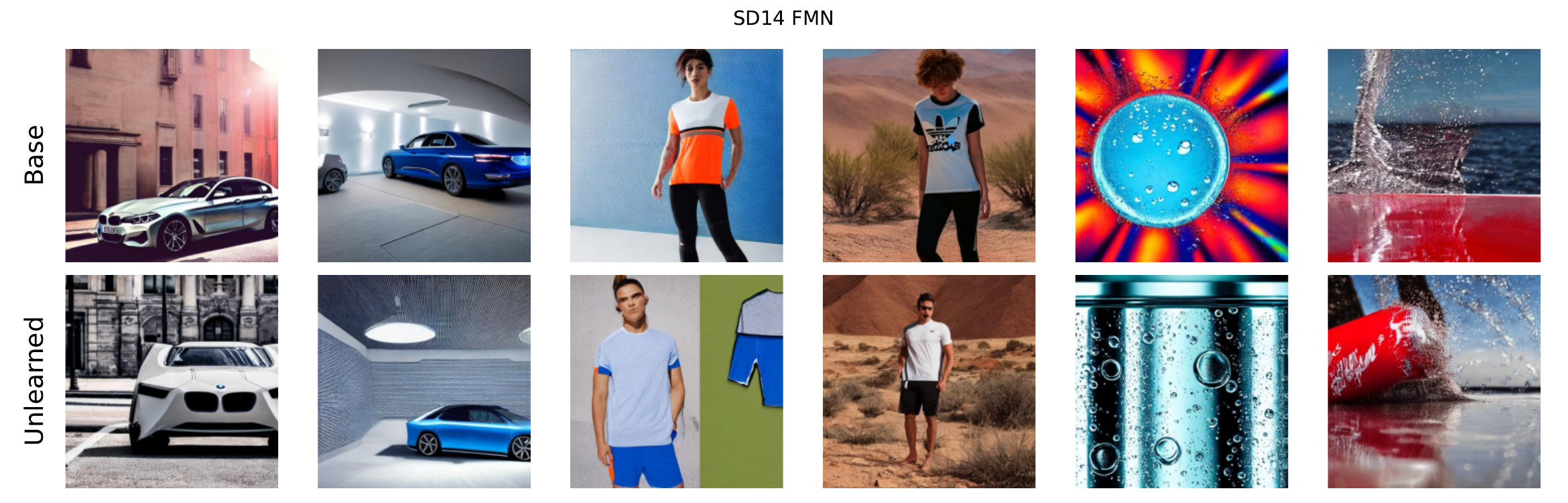}
    
    \label{fig_un_3}
\caption{Evaluation of different machine unlearning methods applied to the unbranding task.}
\label{fig_un_3}
\end{figure}

\begin{figure}
    \centering
    \includegraphics[width=\linewidth]{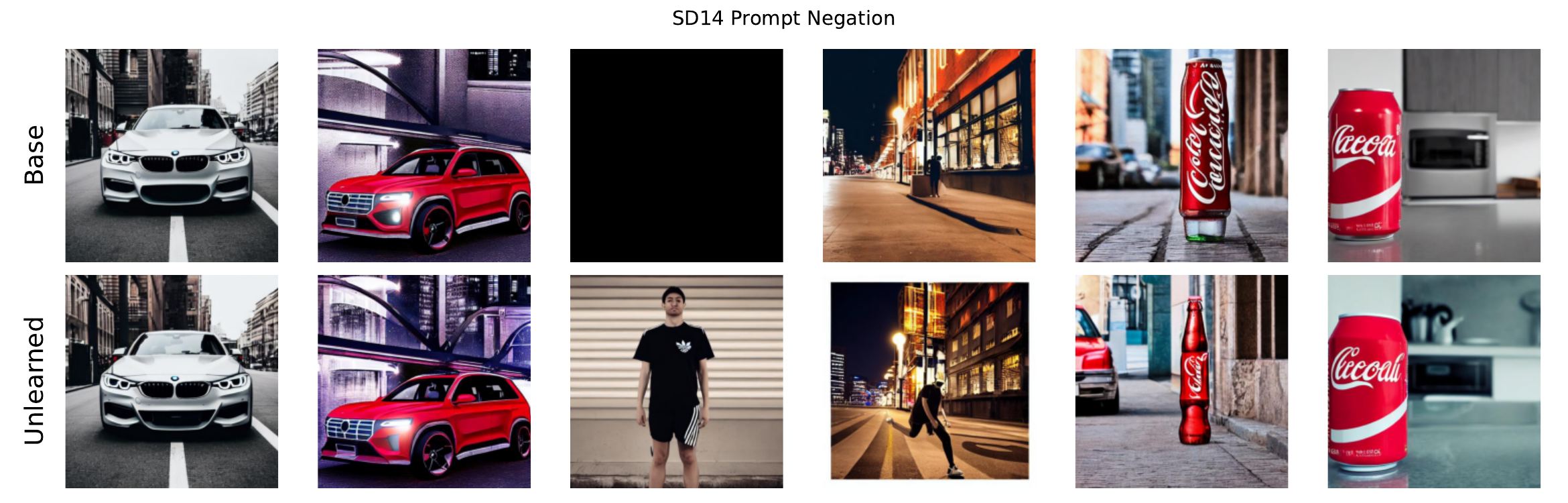}
    \includegraphics[width=\linewidth]{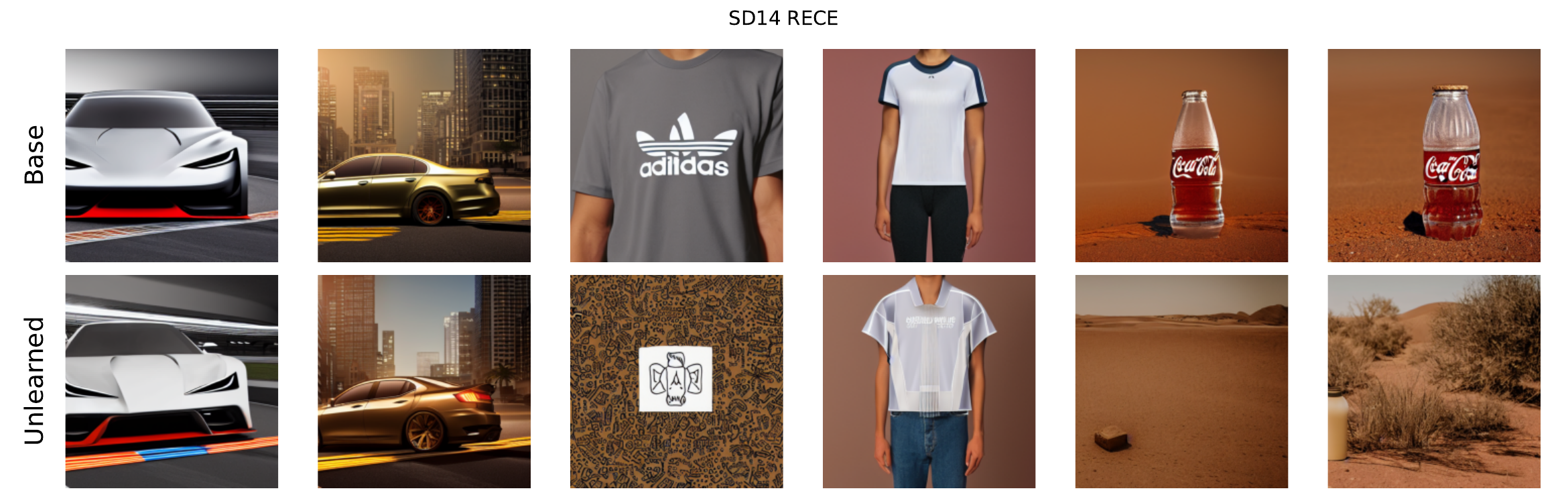}
    \includegraphics[width=\linewidth]{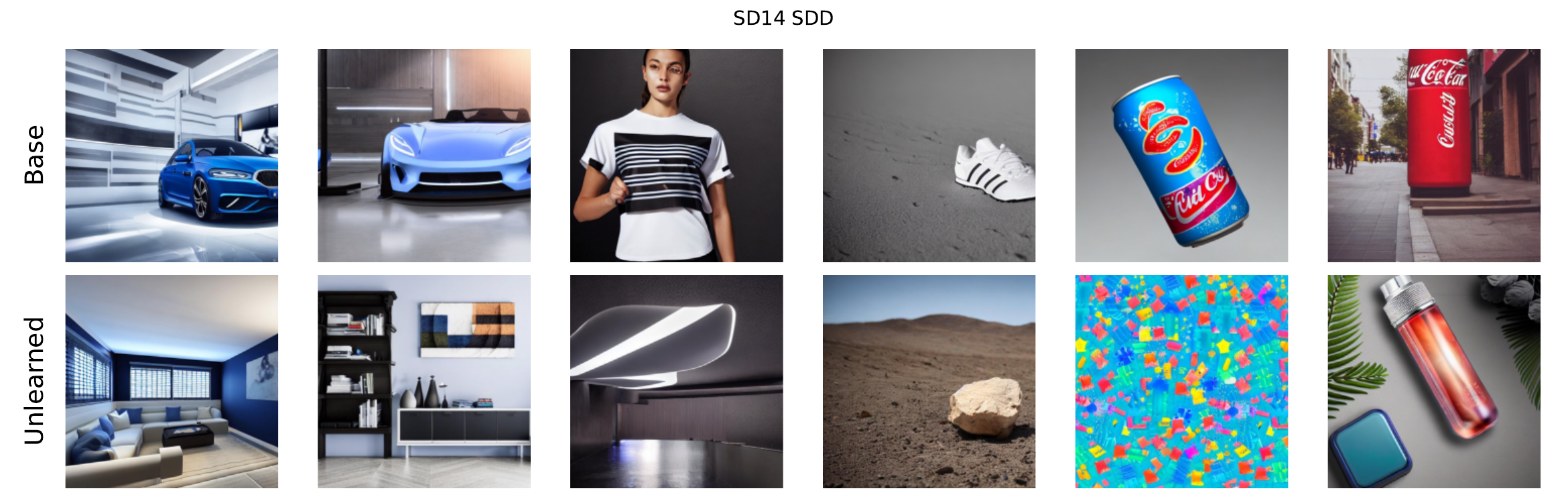}
    \caption{Evaluation of different machine unlearning methods applied to the unbranding task.}
    \label{fig_un_4}
\end{figure}

\end{document}